\def\year{2019}\relax
\documentclass[letterpaper]{article} %DO NOT CHANGE THIS
\usepackage{aaai19}  %Required
\usepackage{times}
\usepackage{epsfig}
\usepackage{graphicx}
\usepackage{amsmath}
\usepackage{amssymb}
\usepackage{xspace}
\usepackage{booktabs}
\usepackage{multirow}
\usepackage{subcaption}
\usepackage{url}
\usepackage{xcolor}
\usepackage[normalem]{ulem} % [normalem] is added as it redefines \em/\emph
\usepackage{verbatim}
\usepackage[font=small]{caption}
\usepackage[ruled,vlined,linesnumbered]{algorithm2e} 
\usepackage{tikz}
\usepackage{wrapfig}
\usepackage{sidecap}
\usepackage{adjustbox}
\usetikzlibrary{decorations.pathreplacing,calc}
\usepackage{color, colortbl,xcolor}
\definecolor{Gray}{gray}{0.9}
\definecolor{LightCyan}{rgb}{0.88,1,1}

\usepackage{microtype}

\newtheorem{theorem}{Theorem}

\title{Deep Domain Adaptation under Deep Label Scarcity}
\author{ Amar Prakash Azad,  Dinesh Garg,  Priyanka Agrawal, Arun Kumar\\
  IBM Research AI\\
  {\tt \{amarazad,garg.dinesh,priyanka.agrawal,kkarun\}@in.ibm.com}
}

\nocopyright
\begin{document}
\setlength\titlebox{4.2cm}

\maketitle
%
%\input{abs.tex}
%%%%%%%%%%%%%%%%%%%%%%%%%%%%%%%%%%%%%%%%%%%%%%%%%%%%%%%%%%%%%%%%%%%%%%%%%%%%%%%
\begin{abstract}
%The goal behind {\em Domain Adaptation (DA)} is to leverage the labeled examples from a source domain so as to infer an accurate model in a target domain. The problem becomes challenging because labels in the target domain are not available or are in scarcity at the best. An state-of-the-art approach for the DA probles is due to \cite{GaninUAGLLML16}, known as {\em DANN}, where they try to find a common representation of source and target domains via adversarial training. This approach requires a large number of labeled examples from the source domain to be able to infer a good model for the target domain. In many situations, however, obtaining labels in the source domain is expensive and which deteriorates the performance of DANN. This limits the applicability of DANN in such scenarios. In this paper, we propose an improved approach called as {\em TransDANN}. Our approach is motivated by the ideas from the transductive learning field. In our approach, first we argue that DANN reduces the original DA problem into a semi-supervised learning problem over the space of common representation. Next, we propose a transductive learning based extension to the DANN training procedure using this reduction. Experimental results (both on text and images) show upto {\bf 11\%} boost in the performance of TransDANN over DANN under deep source label scarcity scenarios.
The goal behind {\em Domain Adaptation (DA)} is to leverage the labeled examples from a source domain so as to infer an accurate model in a target domain where labels are not available or in scarce at the best. A state-of-the-art approach for the DA  is due to \cite{GaninUAGLLML16}, known as {\em DANN}, where they attempt to induce a common representation of source and target domains via adversarial training. 
This approach requires a large number of labeled examples from the source domain to be able to infer a good model for the target domain. 
However, in many situations obtaining labels in the source domain is expensive which results in deteriorated performance of DANN 
%demands DA model in limited source label supply. In these scenarios, performance of DANN  deteriorates the performance of DANN. 
and limits its applicability in such scenarios. In this paper, we propose a novel approach to overcome this limitation. %called as {\em TransDANN}.  leveraging ideas from transductive learning theory.
%In this paper, we propose an improved approach called as {\em TransDANN}. 
%Our approach is motivated by the ideas from the transductive learning theory.  
In our work, we first establish that DANN reduces the original DA problem into a semi-supervised learning problem over the space of common representation. Next, we propose a learning approach, namely {\em TransDANN}, that amalgamates adversarial learning and transductive learning to mitigate the detrimental impact of limited source labels and yields improved performance.
%Next, we propose a transductive learning based extension to the DANN training procedure using this reduction. 
Experimental results (both on text and images) show a significant boost in the performance of TransDANN over DANN under such scenarios. 
We also provide theoretical justification for the performance boost. 
%deep label scarcity scenarios where labels are in scarce in source and unavailable in target.
%Experimental results (both on text and images) show upto {\bf 11\%}  boost in the performance of TransDANN over DANN under such deep source label scarcity scenarios.

\end{abstract}
%%%%%%%%%%%%%%%%%%%%%%%%%%%%%%%%%%%%%%%%%%%%%%%%%%%%%%%%%%%%%%%%%%%%%%%%%%%%

%%%
%\input{intro_arxiv.tex}
%
%%%%%%%%%%%%%%%%%%%%%%%%%%%%%%%%%%%%%%%%%%%%%%%%%%%%%%%%%%%%%%%%%%%%%%%%%%%%
\section{Introduction}
In many real life scenarios, label acquisition %of the data 
is a daunting task due to various limitations including cost, time, hazards, confidentiality, scale, etc. This limits the applicability of many successful machine learning and deep learning models which otherwise require a large number of labeled data. The field of {\em domain adaptation (DA)} aims at easing out learner's job under such stress situations by allowing a transfer of learned models to other domain that faces label scarcity or absence. 
% leverage data for training deep models
An example of such scenarios is commonly observed when immense amount of annotated labelled data \cite{SunS14,Vazquez14}  are created in the source domain whereas the target domain (often real world application domain) lacks annotation. The dissimilarity in marginal distribution of source domain and target domain data, called as {\em covariate shift},  is often significant and detrimental to the performance of source trained model on target data \cite{SHIMODAIRA2000227}.   
On the other hand, the dissimilarity of conditional distribution of source domain and target domain data, {\em concept shift}, can also impact performance of source trained model on target data despite absence of covariate shift \cite{datashiftbook}. 
For example, we might have an email spam filter trained from a large email collection received by a group of current users (the source domain) and wish to adapt it for a new user (the target domain) where we hardly have any email marked as spam by this new user~\cite{Ben-David2010}. A similar situation arises during cold-start of an on-line recommender system when a new customer joins. In the email example, 
intuition suggests that we should be able to improve the performance of spam filter for the new user as long as we believe that users behave consistently in terms of labeling emails as spam or ham, denoted by $P(spam \mid email)$. The challenge, however, is that each user receives a unique distribution of emails, say $P(email\mid user)$. The situation in the recommendation system example, on the other hand, could be little more complex. In this case, the behaviors of the users toward products need not be the same. That is $P(liking \mid product)$ may be different for different users. Furthermore, each user has a unique distribution, denoted by $P(product \mid user)$, from which he browses the products in the catalog. Therefore, transferring the learning may be bit hard. The email problem falls in a category of the DA problems where we say that {\em covariate shift} assumption holds. The recommendation engine problem, on the other hand, falls in the category of the DA problems where we say that both {\em covariate shift} as well as {\em concept shift} are present. 

\cite{Ben-David2010} studied the class of DA problem %where {\em covariate shift} is 
where both {\em covariate shift} and {\em concept shift} are 
present but under the assumption that there exists a labeling rule, say $h^*(\cdot)$, which works good for both the domains. In the very same setting, \cite{GaninUAGLLML16} proposed {\em Domain Adversarial Neural Networks (DANN)} approach which extracts such an $h^*(\cdot)$ in deep learning framework.
%by means of constraining the classifier that rely only on domain-invariant features.
 They achieve this objective  by training the classifier to perform well on the source domain while minimizing the divergence between features extracted from the source versus target domains. For divergence minimization, they used domain adversarial training which leverages the target domain data without the need for their label. The deep learning framework enables to build the mapping between source domain and target domain through the domain classifier of the adversarial training. 
 
As mentioned in \cite{GaninUAGLLML16}, DANN doesn't require labeled examples from target domain but it requires a large number of labeled examples from source domain in order to output a classifier $h(\cdot)$ that is reasonably close to $h^*(\cdot)$. The performance of DANN gets adversely affected when the supply of source domain labelled examples are limited - a situation common in real life. This happens because the error bound given in \cite{Ben-David2010} becomes noisy when labels are less and DANN tries to minimize this bound.

In this paper, we propose a novel approach, called as {\em TransDANN}, by fusing transductive learning theory with adversarial domain adaptation  which prevents DANN suffering from low performance during deep scarcity of source labels.  
{\em TransDANN}  is inspired by an early work of \cite{Joachims:1999} on {\em transductive learning}. 
%Our approach is based on the observation that 
We argue that 
DANN attempts to reduce an original DA problem into a semi-supervised learning problem over the extracted common space of domain-invariant features. This enables one to employ semi-supervised learning techniques for performance boosting. Experimental results (both on text and images) confirm the superiority of TransDANN over DANN. 

%%%%%%%%%%%%%%%
%%% 								Prior Art
%%%%%%%%%%%%%%%%%%%%%%%%%%%%%%%%%%%%%%%%%%%%%%%%%%%%%%%%%%%%%%%%%%%%%%%%%%%%%%%
\subsubsection{Prior Art}
%Given an extensive literature on the problem of {\em domain adaptation} in particular, and {\em transfer learning} in general, we highlight the works that are the most relevant to this paper. 
The survey articles \cite{PatelGLC15}, \cite{Csurka2017}, \cite{WANG2018135} provide a landscape of the DA problem area. 
%Although these survey articles are biased towards the visual domain adaptation but still they can be considered as comprehensive becausethe  majority of the work in the DA focuses on visual domain adaptation. 
 Broadly speaking, DA approaches belong to two categories - (i) {\em conservative} and (ii) {\em non-conservative}. In a conservative approach, information contained in unlabeled examples from the target domain is not leveraged. Whereas, in a non-conservative approach, it is leverage. Theoretical analysis of the conservative approaches can be found in \cite{Ben-David:2006}, \cite{Blitzer2007}, \cite{MansourMR09}. Among non-conservative approaches, one idea is to re-weight the source labeled examples so as to match the marginal distributions of both the domains. \cite{pmlr-v9-david10a} and \cite{Ben-David2010} provided sound theoretical analysis for non-conservative approaches and proved an inevitable bound on the error of the learned hypothesis for the target domain. 
%In fact, the bound presented in \cite{Ben-David2010} paved the way for more recent algorithms in the area of non-conservative domain adaptation whose aim is to match this bound. 
Recent approaches for non-conservative DA are inspired by the recent progress in the areas of deep neural networks and deep generative models \cite{GAN14}. The prominent approach along these lines include \cite{Long2015}, \cite{GaninUAGLLML16}, and \cite{Tzeng2017}. The idea in \cite{GaninUAGLLML16} is to project both source and target marginal distributions into a common feature space and encourage projected distributions to match. They used the idea of generative adversarial nets \cite{GAN14} for this purpose. Other recent works along similar lines include \cite{Saito2017} and \cite{Shu2018ADA}. 
%\cite{Saito2017} proposed to modify tri-training scheme of \cite{Zhou2005} for the context of domain adaptation. 
The approach proposed in \cite{Shu2018ADA} tries to improve DANN under the scenario where clustering assumption holds true for the target domain. 
In the text domain,  \cite{Liu2017,multinomial} used adversarial training to obtain better  generalization through multitask setting where both source and target domain data is available.  
 %- {\em that is, the feature vectors in the target domain exhibit natural clusters and the class boundaries never cross high density regions.} 
 
 To the best of our knowledge, there is no other work which addresses the issue of DANN's performance under source label scarcity. Our work is the first one to identify and address this gap. %We will make the source code also available.
%%%%%%%%%%%%%%%%%%%%%%%%%%%%%%%%%%%%%%%%%%%%%%%%%%%%%%%%%%%%%%%%%%%%%%%%%%%%%%%%%
%                     Background -- Domain Adaptation Problem
%%%%%%%%%%%%%%%%%%%%%%%%%%%%%%%%%%%%%%%%%%%%%%%%%%%%%%%%%%%%%%%%%%%%%%%%%%%%%%%%%

\subsubsection{Background -- DA Problem Setup}
%As suggested by \cite{Ben-David2010}, we define 
A domain $\cal{D}$ is defined as a tuple ${\cal{D}}=\left< {\cal{X}}, {\cal{Y}}, {\mathbb{P}}({\mathbf{x}}, y)\right>$, where $\cal{X}$ denotes the feature space, $\cal{Y}$ denotes the label space, and ${\mathbb{P}}({\mathbf{x}}, y)$ denotes the joint probability distribution function over the space ${\cal{X}} \times {\cal{Y}}$.

In a typical DA problem setup, we are given a source domain ${\cal{D}}_s=\left< {\cal{X}}_s, {\cal{Y}}_s, {\mathbb{P}}_s({\mathbf{x}}_s, y_s)\right>$ and a target domain ${\cal{D}}_t=\left< {\cal{X}}_t, {\cal{Y}}_t, {\mathbb{P}}_t({\mathbf{x}}_t, y_t)\right>$. The Bayes theorem allows us to write the density functions of the source and the target distributions as follows:~\footnote{We use symbol ${\mathbb{P}}(\cdot)$ to denote a distribution function and $P(\cdot)$ to denote the corresponding density function.}
${{P}}_d({\mathbf{x}}_d, y_d) = {{P}}_d(y_d \mid {\mathbf{x}}_d) \cdot {{P}}_d({\mathbf{x}}_d),\; \text{where } d\in\{s,t\}$. The density functions ${{P}}_s(y_s \mid {\mathbf{x}}_s) $ and ${{P}}_t(y_t \mid {\mathbf{x}}_t)$ are typically referred to as {\em conditionals}, whereas the functions ${{P}}_s({\mathbf{x}}_s) $ and ${{P}}_t({\mathbf{x}}_t)$ are referred to as {\em marginals}. In this paper, we assume ${\cal{Y}}_s={\cal{Y}}_t={\cal{Y}} =\{0,1\}$.
%, given by $\{1,2,\ldots,\ell\}$. 
However, our results are applicable as long as ${\cal{Y}}_s={\cal{Y}}_t={\cal{Y}}$ and ${\cal{Y}}$ is any other label space which can be handled by deep neural networks. 
% This assumption would mean that the source and the target domains can be given by the respective probability distributions ${\mathbb{P}}_s({\mathbf{x}}_s, y_s)$ and ${\mathbb{P}}_t({\mathbf{x}}_t, y_t)$ defined over the joint spaces ${\cal{X}}_s \times {\cal{Y}}$ and ${\cal{X}}_t \times {\cal{Y}}$, respectively. These joint distributions are typically called as the {\em source} and the {\em target} distributions, respectively. 

The goal of any DA problem is to predict the label $y_t \in \cal{Y}$ for any given target sample ${\mathbf{x}}_t \in {\cal{X}}_t$ drawn from ${\mathbb{P}}_t({\mathbf{x}}_t)$. The assumption is that both ${\mathbb{P}}_s({\mathbf{x}}_s, y_s)$ and ${\mathbb{P}}_t({\mathbf{x}}_t, y_t)$ are unknown to the learner. The only information available with the learner at the time of training is labeled examples (say $n$) from the {\em source domain} and {unlabeled} examples from the {\em target domain}, say $N$. We denote these training data by $D_s=\{({\mathbf{x}}_{s}^i, y_s^i)\}_{i=1}^n$, and $D_t=\{{\mathbf{x}}_{t}^j\}_{j=1}^N$, respectively. 
%We also denote the unlabeled examples from the source domain as ${{U}}_s=\{{\mathbf{x}}_{s}^i\}_{i=1}^{n}$. For the consistency sake, we also use an alternative symbol ${{U}}_t$ to denote the unlabeled examples from the target domain, that is $U_t=D_t$   

Most of the DA work hinges around the assumption of ${\cal{X}}_s={\cal{X}}_t={\cal{X}}$ and this setting is known as {\em homogeneous DA}~\cite{WANG2018135}. For the binary classification problem $\left({\cal{Y}} =\{0,1\}\right)$ in this setting, \cite{Ben-David2010} gave a result (stated below) that relates the accuracy of any labeling function (aka hypothesis) $h: {\cal{X}} \mapsto \{0,1\}$ on the source domain with the accuracy of the same hypothesis on the target domain. 
%In this theorem, we have done away with the subscript $s$ and $t$ due to the homogeneous setting assumption. 
\begin{theorem}[\cite{Ben-David2010}]\label{Ben-David_Theorem1}
Let $\cal{H}$ be a hypothesis space of VC dimension $d$, then for any $\delta \in (0,1)$, with probability $1-\delta$ (over the choice of samples), for every $h \in {\cal{H}}$:
$\epsilon_s(h) \le \epsilon_t(h) + \frac{1}{2} {{d}}_{{\cal{H}} \Delta {\cal{H}}}({{\mathbb{P}}_s}(\mathbf{x}), {{\mathbb{P}}_t}(\mathbf{x}))+\lambda$
% \begin{eqnarray*}
% \epsilon_s(h) \le \epsilon_t(h) + \frac{1}{2} {{d}}_{{\cal{H}} \Delta {\cal{H}}}({{\mathbb{P}}_s}(\mathbf{x}), {{\mathbb{P}}_t}(\mathbf{x}))+\lambda
% \end{eqnarray*} 
\end{theorem}
where, error $\epsilon_s(h)$ and $\epsilon_t(h)$ are defined as the expected loss for the hypothesis $h$ with respect to the source and the target domain's conditional distribution, respectively. That is, 
$\epsilon_d(h) = \mathbb{E}_{ {\mathbf{x}} \sim {{\mathbb{P}}_d}(\cdot)}\left[\;\left|h(\mathbf{x}) - {\mathbb{P}}_d(y=1 \mid {\mathbf{x}})\right|\;\right];\text{where } d \in\{s,t\}$.
% Similarly, the error $\epsilon_t(h)$ is given as follows.
% \begin{eqnarray}
% \epsilon_t(h) = \mathbb{E}_{{\mathbf{x}} \sim {{\mathbb{P}}_t}(\cdot)}\left[\;\left|h(\mathbf{x}) - {\mathbb{P}}_t(y=1 \mid {\mathbf{x}})\right|\;\right] 
% \end{eqnarray}
The quantity ${{d}}_{{\cal{H}} \Delta {\cal{H}}}({{\mathbb{P}}_s}(\mathbf{x}), {{\mathbb{P}}_t}(\mathbf{x}))$ denotes the ${\cal{H}} \Delta {\cal{H}}$ distance between the distributions ${\mathbb{P}}_s({\mathbf{x}})$ and ${\mathbb{P}}_t({\mathbf{x}})$ and it  
%and is widely used in the theory of non-conservative domain adaptation. 
%This quantity 
accounts for the gap between $\epsilon_s(h)$ and $\epsilon_t(h)$ arising due to the discrepancy between ${\mathbb{P}}_s({\mathbf{x}})$ and ${\mathbb{P}}_t({\mathbf{x}})$. This quantity is given by the following expression:
${{d}}_{{\cal{H}} \Delta {\cal{H}}}({{\mathbb{P}}_s}(\mathbf{x}), {{\mathbb{P}}_t}(\mathbf{x})) = \underset{h \in {\cal{H}} \Delta {\cal{H}}}{2\sup} \left[ \left| \alpha(h)-1\right|\right]$
%&\underset{h \in {\cal{H}} \Delta {\cal{H}}}{2\sup} \left[\left|P_{\mathbf{x} \sim {\mathbb{P}}_s(\mathbf{x})}\left(h(\mathbf{x})=1\right)- P_{\mathbf{x} \sim {\mathbb{P}}_t(\mathbf{x})}\left(h(\mathbf{x})=1\right)\right|\right] 
% \nonumber\\
% & = \underset{h \in {\cal{H}} \Delta {\cal{H}}}{2\sup} \left[ \left| \alpha(h)-1\right|\right]
where, 
$\alpha(h)=P_{\mathbf{x} \sim {\mathbb{P}}_s(\mathbf{x})}\left\{\mathbf{x}|h(\mathbf{x})=0\right\}+ P_{\mathbf{x} \sim {\mathbb{P}}_t(\mathbf{x})}\left\{\mathbf{x}|h(\mathbf{x})=1\right\}$. 
The ${{\cal{H}} \Delta {\cal{H}}}$ constitutes the space of hypotheses which are pairwise symmetric difference of any two hypotheses from $\cal{H}$. 
%The quantity $\alpha(h)$ can be interpreted as if it were to measure the accuracy of a binary classifier $h:{\cal{X}} \mapsto \{0,1\}$ which classifies the unlabeled examples from the source domain and the target domain. 
By looking at $\alpha(h)$, we can say that ${\cal{H}} \Delta {\cal{H}}$ distance between the marginals of the source domain and the target domain can be calculated by identifying the best domain classifier $h \in {\cal{H}} \Delta {\cal{H}}$ which classifies the unlabeled examples from the source domain and the target domain. The exact same idea was exploited by \cite{GaninUAGLLML16}. The last term $\lambda$ is given by $\lambda = \underset{h \in {\cal{H}}}{\min}\left(\epsilon_s(h)+\epsilon_t(h)\right)$.

Following theorem is a refined version of Theorem \ref{Ben-David_Theorem1} for the scenario when one has an empirical estimate $\hat{\alpha}(h)$ of the quantity ${\alpha}(h)$ (and thereby, an empirical estimate ${\hat{d}}_{{\cal{H}} \Delta {\cal{H}}}(U_s, U_t)$). This empirical estimate can be computed by having access to say $m$ unlabeled examples ${{U}}_s$ and $U_t$ drawn from ${\mathbb{P}}_s({\mathbf{x}})$ and ${\mathbb{P}}_t({\mathbf{x}})$, respectively.   

% It was also shown in \cite{Ben-David2010} that if ${{U}}_s$ and $U_t$ are unlabeled samples of size $m$ each, drawn from the distribution ${\mathbb{P}}_s({\mathbf{x}})$ and ${\mathbb{P}}_t({\mathbf{x}})$, respectively, then one can get an empirical estimate of the quantity ${{d}}_{{\cal{H}} \Delta {\cal{H}}}({{\mathbb{P}}_s}(\mathbf{x}), {{\mathbb{P}}_t}(\mathbf{x}))$ as follows: ${\hat{d}}_{{\cal{H}} \Delta {\cal{H}}}(U_s, U_t)= \underset{h \in {\cal{H}} \Delta {\cal{H}} }{2\sup} \left[\; \left| \hat{\alpha}(h) -1\right| \;\right]$ where, $\hat{\alpha}(h)$ is an empirical estimate of $\alpha(h)$ based on the samples $U_s$ and $U_t$. Furthermore, this empirical estimate relates with the original quantities $\epsilon_s(h)$ and $\epsilon_t(h)$ in the following manner, yielding the main theorem of \cite{Ben-David2010} which is  refined version of Theorem \ref{Ben-David_Theorem1} above.
%%%%%%%%%%%%%%%%%%%%%
% \begin{lemma}[\cite{Ben-David2010}]
% \begin{eqnarray*}
% {{d}}_{{\cal{H}} \Delta {\cal{H}}}({{\mathbb{P}}_s}(\mathbf{x}), {{\mathbb{P}}_t}(\mathbf{x})) &\le&
% {\hat{d}}_{{\cal{H}} \Delta {\cal{H}}}(U_s, U_t) \nonumber \\
% &&+ 4 \sqrt{\frac{d \log(2m) + \log(\frac{2}{\delta})}{m}}
% \end{eqnarray*}
% \end{lemma}
% Combining this lemma, with above theorem, resulted in the main theorem of \cite{Ben-David2010}
%%%%%%%%%%%%%%%%%%%%%
\begin{theorem}[\cite{Ben-David2010}] \label{ben-david-theorem}
% Let $\cal{H}$ be a hypothesis space of VC dimension $d$. If ${{U}}_s$ and $U_t$ are unlabeled samples of size $m$ each, drawn from the distribution ${\mathbb{P}}_s({\mathbf{x}})$ and ${\mathbb{P}}_t({\mathbf{x}})$, respectively, then for any $\delta \in (0,1)$, with probability $1-\delta$ (over the choice of samples), for every $h \in {\cal{H}}$:
\begin{eqnarray*}
&\epsilon_s(h) \le \epsilon_t(h) +\frac{{\hat{d}}_{{\cal{H}} \Delta {\cal{H}}}(U_s, U_t) }{2}+\sqrt{\frac{d \log(2m) + \log(\frac{2}{\delta})}{m/16}}+\lambda
\end{eqnarray*}
\end{theorem}
This theorem offers the following insight. {\em In order to find a good hypothesis $h(\cdot)$ for the target domain, one should aim to find a hypothesis space $\cal{H}$ that not only contains a good hypothesis $h \in \cal{H}$ for the source domain, but also the best domain classifier in the space ${{\cal{H}} \Delta {\cal{H}}}$ is as poor as possible.}
%%%%%%%%%%%%%%%%%%%%%%%%%%%%%%%%%%%%%%%%%%%%%%%%%%%%%%%%%%%%%%%%%%%%%%%%%%%%%%%%%
%                 Domain Adversarial Neural Networks (DANN)
%%%%%%%%%%%%%%%%%%%%%%%%%%%%%%%%%%%%%%%%%%%%%%%%%%%%%%%%%%%%%%%%%%%%%%%%%%%%%%%%%

%
%
%\input{DANN.tex}
%
\section{Domain Adversarial Neural Networks (DANN)}
Motivated by the above insights, 
%inferred from the theoretical result of \cite{Ben-David2010}, 
\cite{GaninUAGLLML16} proposed a novel feedforward neural network architecture, known as Domain Adversarial Neural Networks (DANN). 
%The goal behind training the DANN is to find a hypothesis space $\cal{H}$ which not only contains a hypothesis $h \in \cal{H}$ that is good for the source domain, but also the classification accuracy of the best domain classifier picked from the corresponding space ${{\cal{H}} \Delta {\cal{H}}}$ is as low as possible. 
%The DANN is pretty much the state-of-the-art. 
%In the same stroke, the ADA approach of \cite{GaninUAGLLML16} also relaxed the assumption of ${\cal{X}}_s = {\cal{X}}_t={\cal{X}}$ used in Theorem \ref{ben-david-theorem}. 

The DANN architecture starts with a mapping, $G_f: {\cal{X}} \mapsto {\mathbb{R}}^{d}$, called as {\em feature map}, parameterized by the parameter $\theta_f$. This feature map essentially projects any given unlabeled (source or target) example into a $d$-dimensional Euclidean feature space. These feature vectors are then mapped to the class label (more generally, $P(y=1\mid \mathbf{f})$) by means of another mapping, $G_y: {\mathbb{R}}^{d} \mapsto [0,1]$, called as {\em label predictor}. Lastly, the same feature vector $\mathbf{f}$ is mapped to the domain label by means of the mapping, $G_d: {\mathbb{R}}^{d} \mapsto [0,1]$, known as {\em domain classifier}. The respective parameters of the label predictor and domain classifier are denoted by $\theta_y$ and $\theta_d$, respectively. The hypothesis space $\cal{H}$ for this network becomes the composition of $G_y$ and $G_f$, given by ${\cal{H}}=\{h\mid h(\mathbf{x};\theta_f, \theta_y) = G_y(G_f(\mathbf{x}; \theta_f); \theta_y)\}$, and the symmetric difference space becomes ${{\cal{H}} \Delta {\cal{H}}}=\{h\mid h(\mathbf{x};\theta_f, \theta_d) = G_d(G_f(\mathbf{x}; \theta_f); \theta_d)\}$. 

The training of DANN is very interesting. Note, the parameter $\theta_f$ is common to both hypothesis space $\cal{H}$ as well as the symmetric difference hypothesis space ${{\cal{H}} \Delta {\cal{H}}}$. It is this parameter $\theta_f$ which 
on the one hand (along with the parameters $\theta_y$) helps tuning $\cal{H}$ so as to include a hypothesis $h$ of low source domain error $\epsilon_s(h)$ (first term in Equation (\ref{loss_func})). While, on the other hand, it helps (along with the parameters $\theta_d$) adjusting the space ${{\cal{H}} \Delta {\cal{H}}}$ so as the best domain classifying hypothesis $h'\in {{\cal{H}} \Delta {\cal{H}}}$ becomes as poor as possible (last two terms in Equation (\ref{loss_func})). This is achieved by finding the saddle point of the following loss function: ${\cal{L}}(\theta_f,\theta_y,\theta_d,D_s,D_t) =$                
\begin{eqnarray}
%& {\cal{L}}(\theta_f,\theta_y,\theta_d,D_s,D_t) = 
&\sum\nolimits_{i=1}^{n} \frac{L(h({\mathbf{x}}_s^{i};\theta_f, \theta_y), y_s^i)}{n} - \lambda \left[\sum\nolimits_{i=1}^{n} \frac{L(h({\mathbf{x}}_s^{i};\theta_f, \theta_d), d_s^i=0)}{n} + 
\right.\nonumber \\
&
%\right. \nonumber\\ && \left. 
\left. \sum\nolimits_{j=1}^{N} \frac{L(h({\mathbf{x}}_t^{j};\theta_f, \theta_d), d_t^j=1)}{N}\right] \label{loss_func}
\end{eqnarray}
%\begin{eqnarray}
% &\hspace{-1.5cm}{\cal{L}}(\theta_f,\theta_y,\theta_d,D_s,D_t) = 
%\sum\nolimits_{i=1}^{n} \frac{L(h({\mathbf{x}}_s^{i};\theta_f, \theta_y), y_s^i)}{n} - \nonumber\\
%&\hspace{-0.5cm} \lambda \left[\sum\nolimits_{i=1}^{n} \frac{L(h({\mathbf{x}}_s^{i};\theta_f, \theta_d), %d_s^i=0)}{n} + 
%\left. \sum\nolimits_{j=1}^{N} \frac{L(h({\mathbf{x}}_t^{j};\theta_f, \theta_d), d_t^j=1)}{N}\right] %\label{loss_func}
%\end{eqnarray}
where, $h({\mathbf{x}};\theta_f, \theta_y)= G_y(G_f({\mathbf{x}},\theta_f);\theta_y)$, $h({\mathbf{x}};\theta_f, \theta_d)= G_d(G_f({\mathbf{x}},\theta_f);\theta_d)$, $\lambda$ is a hyper-parameter, and $L(\cdot, \cdot)$ is a cross-entropy loss function.
%negative log-likelihood loss function. If true label is $y\in \{0,1\}$ and the predicted score $s$ is given by $s=p(y=1\mid \mathbf{x})$, then one would have  $L(s,y) = - y \log(s) - (1-y) \log(1-s)$.
% \begin{eqnarray*}
% L(s,y) = - y \log(s) - (1-y) \log(1-s) 
% \end{eqnarray*} 
The labels $d_s^i$ and $d_t^j$ represent the labels to identify the domain (source being $0$ and target being $1$) of any training example $\mathbf{x}$. The training of DANN proceeds in iterations where the aim is to find the saddle point $\hat{\theta}_f,\hat{\theta}_y,\hat{\theta}_d$ of ${\cal{L}}(\theta_f,\theta_y,\theta_d,D_s,D_t)$ defined as below.
\begin{eqnarray}
\hat{\theta}_f,\hat{\theta}_y &=& \underset{{\theta}_f,{\theta}_y}{\arg\min}\; {\cal{L}}(\theta_f,\theta_y,\hat{\theta}_d,D_s,D_t) \label{saddle_eqn_1}\\
\hat{\theta}_d &=& \underset{{\theta}_d}{\arg\max} \;{\cal{L}}(\hat{\theta}_f,\hat{\theta}_y,\theta_d,D_s,D_t) \label{saddle_eqn_2}
\end{eqnarray}

The training of DANN is known to require {\em a large amount of labeled} examples from the source domain and {\em a large amount of unlabeled} examples from the target domain. 
%To make this paper self contained, in what follows, we present a quick overview of DANN because 
Our proposed approach improves upon its training under the realistic setting when {\em labeled examples in the source domain are few in numbers}.
%%%%%%%%%%%%%%%%%%%%%%%%%%%%%%%%%%%%%%%%%%%%%%%%%%%%%%%%%%%%%%%%%%%%%%%%%%%%%%%%%
%  				        The Problem of Label Scarcity
%%%%%%%%%%%%%%%%%%%%%%%%%%%%%%%%%%%%%%%%%%%%%%%%%%%%%%%%%%%%%%%%%%%%%%%%%%%%%%%%%
\subsection{The Problem of Label Scarcity}
\begin{figure}[h]
%\hspace*{0.2cm}
\centering
%{\includegraphics[scale=0.45]{images/DANN_Detoriation_3l}}
{\includegraphics[height=0.4\linewidth, width =1\linewidth]{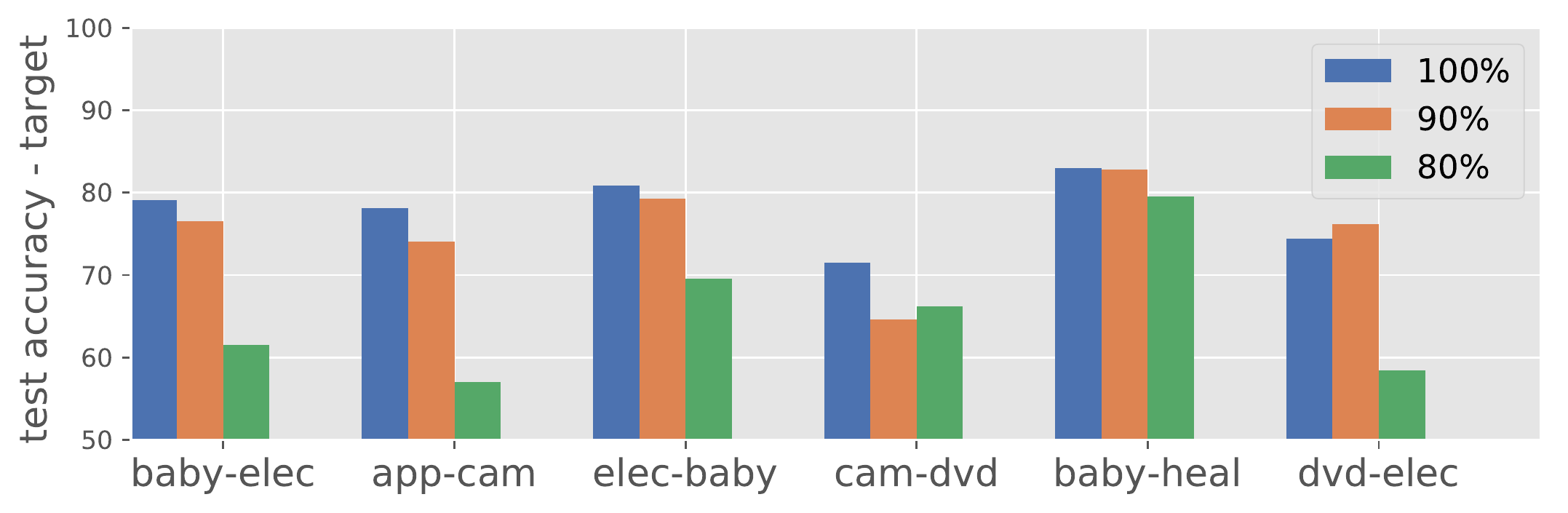}}

%\vspace*{-0.3cm}
\caption{DANN Performance with Reduced Label Supply}
\label{fig:DANN_Detoriation}
\vspace*{-0.23cm}
\end{figure}

% Motivated by the above insights, \cite{GaninUAGLLML16} proposed a novel feed forward neural network architecture (known as {\em Domain Adversarial Neural Networks (DANN)}) for finding such a hypothesis space $\cal{H}$. This approach of DANN is pretty much the state-of-the-art as of today. The training of DANN requires {\em large amount of labeled} examples from the source domain and {\em large amount of unlabeled} examples from the target domain. 
%In many situations, acquiring the large number of labels for the source examples is prohibitive due to various reasons -- cost, time, hazards, confidentiality, etc. The immediate question that surfaces out in such a situation is whether DANN can still be trained effectively. 
Our initial experiments suggest (see Figure \ref{fig:DANN_Detoriation}) that as we shrink the supply of labeled examples during DANN training, the resulting hypothesis of DANN deteriorates. In this figure, we have shown the performance of 6 different (source,target) domain pairs from Amazon product review dataset. For each pair, we have compared the performance of DANN output as we reduce the supply of labeled examples from 100\% to 80\%. The reasons behind such a behavior could be as follows. Given that DANN aims to minimize the error bound of Theorem \ref{ben-david-theorem}, the estimate of the first term in this bound becomes noisy under label scarcity. This motivates us to revisit this problem and investigate whether one can improve the DANN training so as to handle {\em label scarcity.} 
%Before we get to the proposed solution, let us first offer more compelling arguments in support of the observations depicted through Figure \ref{fig:DANN_Detoriation}.
%Imagine a situation where labeled source examples are few in numbers but the unlabeled examples from the target domain are large in numbers. That is, $n$ is very small but $N$ is very large. Additionally, there could be a large number of unlabeled examples from the source domain as well. In a situation like this, one can verify by looking at the loss function ${\cal{L}}(\theta_f,\theta_y,\theta_d,D_s,D_t)$, given in Equation (\ref{loss_func}), that the first term would become quite insignificant relative to the second and the third terms. Therefore, with high probability, training of DANN would spit out a feature map $G_f(\cdot)$ and a label predictor $G_y(\cdot)$ such that the resulting hypothesis $h(\mathbf{x};\hat{\theta}_f,\hat{\theta}_y)= G_y(G_f(\mathbf{x};\hat{\theta}_f);\hat{\theta}_y)$ would perform poorly when predicting the labels of the target domain. The reason being that such a hypothesis would mostly not be a good hypothesis even for the source domain in the first place, let alone the target domain, and therefore, the bound in Theorem \ref{ben-david-theorem}, due to \cite{Ben-David2010}, would kick in.
The key contribution of this paper lies in improving the DANN training so as to handle the deep label scarcity in the source domain. Our idea is based on the two key observations. %which we are going to elaborate upon in the subsequent sections.  
\begin{enumerate}
\item The training of DANN with a large amount of unlabeled examples (both source and target domain) reduces the original DA problem into a semi-supervised learning problem over the single domain of common feature space.
\item The resulting semi-supervised learning problem can be tackled in a way similar to the transductive learning problem handled in \cite{Joachims:1999}.   	
\end{enumerate}
%%%%%%%%%%%%%%%%%%%%%%%%%%%%%%%%%%%%%%%%%%%%%%%%%%%%%%%%%%%%%%%%%%%%%%%%%%%%%%%%%
%         Reduction to Semi-Supervised Learning Problem via DANN
%%%%%%%%%%%%%%%%%%%%%%%%%%%%%%%%%%%%%%%%%%%%%%%%%%%%%%%%%%%%%%%%%%%%%%%%%%%%%%%%%
\subsection{Reduction to Semi-Supervised Learning Problem}
Recall,
%the feature map $G_f(\cdot\;;{\hat{\theta}}_f)$ produced by DANN aims to match induced marginal distributions for the source and the target domains in the feature space $\mathbf{f} \in {\mathbb{R}}^d$. Therefore, when the capacity of the network $G_f(\cdot)$ is large enough, a large pool of unlabeled examples from both the source and the target domains would push the domain classifier $G_d(G_f(\mathbf{x},\theta_f),\theta_d)$ in such a way that it can't discriminant the domains of the unlabeled examples when expressed in the feature space ${\mathbb{R}}^d$. That would mean 
that the marginal distributions ${\mathbb{P}}_s(\mathbf{x})$ and ${\mathbb{P}}_t(\mathbf{x})$ when pushed onto the feature space ${\mathbb{R}}^d$ under the map $G_f(\cdot)$ obtained by training a DANN would be nearly identical.
%Therefore, using the feature map produced at the end of DANN training, one can transform the source and the target domains into the feature space $\mathbf{f}$ where both marginal matches and thereby satisfy the first assumption of Transductive SVM setting \cite{Joachims:1999} as mentioned earlier.
%In summary, we can say that the under some reasonable conditions, the mapping $G_f(\cdot)$ obtained by training a DANN would ideally equalize the marginal distributions induced by the source and the target marginals in the feature space. 
% That is, for any $\mathbf{f} \in {\mathbb{R}}^d$,
% \begin{eqnarray}
% P_f(\mathbf{f}) := {{Pr}}_{\mathbf{x} \sim {\mathbb{P}}_s(\mathbf{x})}(\{\mathbf{x} \mid G_f(\mathbf{x}) = \mathbf{f}\}) = {{Pr}}_{\mathbf{x} \sim {\mathbb{P}}_t(\mathbf{x})}(G_f(\mathbf{x})) \label{marginal_matching_eqn}
% \end{eqnarray}
We denote such an induced marginal distribution in the feature space by ${\mathbb{P}}_f(\mathbf{f})$ and its density by $P_f(\mathbf{f})$.   
%Further, a common assumption in the domain adaptation literature is that of {\em covariate shift}~\cite{SHIMODAIRA2000227}. 
Under covariate shift assumption~\cite{SHIMODAIRA2000227}, one has $P_s(y \mid \mathbf{x}) =P_t(y \mid \mathbf{x})=P(y\mid \mathbf{x})$.
% \[ 
% P_s(y \mid \mathbf{x}) =P_t(y \mid \mathbf{x})=P(y\mid \mathbf{x})
% \]
Thus, under the covariate shift assumption, one can use the output feature map $G_f(\cdot;{\hat{\theta}}_f)$ of DANN to transform both source and target spaces ${{\cal{X}}_s}$ and ${{\cal{X}}_t}$ into the feature space $\mathbf{f}$ where the problem now sounds more like a semi-supervised learning problem having few labeled examples and a large number of unlabeled examples. The labels of the examples in this feature space can be assumed to be sampled from some underlying distribution, say ${\mathbb{P}}_f(y \mid \mathbf{f})$, and the feature vectors $\mathbf{f}$ themselves can be assumed to be sampled from the common induced distribution ${\mathbb{P}}_f(\mathbf{f})$. Any parameterized label classifier $P_f(y \mid \mathbf{f})=G_y(\mathbf{f}; \theta_y)$ defined on this feature space can be combined with the feature map to render a classifier for the source or target domain. That is, $P(y\mid \mathbf{x})=G_y(G_f({\mathbf{x}};\theta_f), \theta_y)$. 

Training a classifier $P_f(y \mid \mathbf{f})$ in this feature space, thus, becomes a semi-supervised learning problem in its own. DANN outputs one such classifier given by $P_f(y \mid \mathbf{f})=G_y(\mathbf{f}; {\hat{\theta}}_y)$. Any improvement on top of this classifier would indeed improve the accuracy of the target domain classifier as well. Therefore, we propose to invoke the approach of semi-supervised learning to offer a classifier $P_f(y \mid \mathbf{f})$ that is better than what is DANN offers, especially when source labels are in scarcity.  
%~\footnote{Whenever it is clear from the context, we drop the subscripts $s$ and $t$ from the example ${\mathbf{x}}$ and the label $y$ so as to make the notations less cluttered.}.
%%%%%%%%%%%%%%%%%%%%%%%%%%%%%%%%%%%%%%%%%%%%%%%%%%%%%%%%%%%%%%%%%%%%%%%%%%%%%%%%%
%          Section : Domain Adversarial Neural Networks (DANN)
%%%%%%%%%%%%%%%%%%%%%%%%%%%%%%%%%%%%%%%%%%%%%%%%%%%%%%%%%%%%%%%%%%%%%%%%%%%%%%%%%
% \subsection{What if Labeled Examples are Few in Numbers?}
% In what follows, we highlight an important observations pertaining to the training of DANN. This observation will form the basis of our proposed approach for improving the training of DANN. 

% In what follows, we first summarize the DANN architecture and its training procedure followed by justifying the use of ideas from \cite{Joachims:1999} in order to tackle the label scarcity issue.
%%%%%%%%%%%%%%%%%%%%%%%%%%%%%%%%%%%%%%%%%%%%%%%%%%%%%%%%%%%%%%%%%%%%%%%%%%%%%%%%%
%    				      Transductive Learning Idea
%%%%%%%%%%%%%%%%%%%%%%%%%%%%%%%%%%%%%%%%%%%%%%%%%%%%%%%%%%%%%%%%%%%%%%%%%%%%%%%%%
\subsection{Transductive Learning to Tackle Label Scarcity}
%In this section, we briefly discuss the {\em Transductive SVM} idea of \cite{Joachims:1999} which inspired our idea behind fixing the label scarcity issue in DANN. The idea behind Transductive SVM is as follows.
Let ${\cal{D}}=\left< {\cal{X}}, {\cal{Y}}, {\mathbb{P}}({\mathbf{x}}, y)\right>$ be some domain. We need not confuse this with source (or target) domain being discussed so far. 
%Let us being with writing down the marginal and the conditional densities for the distribution ${\mathbb{P}}({\mathbf{x}}, y)$ as follows: ${{P}}({\mathbf{x}}, y) = {{P}}(y \mid {\mathbf{x}}) {{P}}({\mathbf{x}})$. 
% \begin{eqnarray*}
% {{P}}({\mathbf{x}}, y) = {{P}}(y \mid {\mathbf{x}}) {{P}}({\mathbf{x}})
% \end{eqnarray*}
Suppose, a learner has access to $n$ labeled examples, $S_{\ell}=\{({\mathbf{x}}_i,y_i)\}_{i=1}^{n}$, drawn independently from the distribution ${\mathbb{P}}({\mathbf{x}}, y)$. 
%Denote the set of these labeled examples as $S_{\ell}=\{({\mathbf{x}}_i,y_i)\}_{i=1}^{n}$. 
The learner also has access to a large number of unlabeled examples $S_u=\{{\mathbf{x}}_j\}_{j=1}^{N}$ drawn from the corresponding marginal distribution ${\mathbb{P}}({\mathbf{x}})$. 
%Denote the set of these unlabeled examples as $S_u=\{{\mathbf{x}}_j\}_{j=1}^{N}$. 
The goal of the learner is to pick a hypothesis from the space ${\cal{H}_{\theta}}=\{h(\cdot\;; \theta) \mid h:{\cal{X}} \mapsto \cal{Y}\}$ so as to predict the labels of the examples in the set $S_u$ as accurately as possible. 
% Below are the two key assumptions for this setting.
% \begin{enumerate}
% \item The labeled samples $\{\mathbf{x}_i\}_{i=1}^{n}$ and the unlabeled samples $\{\mathbf{x}_j\}_{j=1}^{N}$ are both drawn from the same marginal distribution ${{P}}({\mathbf{x}})$.
% \item The labeled samples are few in numbers but the unlabeled samples are really large in numbers. That is, $n$ is very small but $N$ is very large.
% \end{enumerate}

%For this setting, \cite{Joachims:1999} considered a linear hypothesis space parameterized by $\boldsymbol{w} = \theta$ and used Support Vector Machine (SVM) approach to pick a good hypothesis. However, one can use any other hypothesis space and loss functions of their choice. 
For this kind of problems, \cite{Joachims:1999} proposed Transductive SVM approach where the idea is to minimize an appropriate loss function so as to find the joint optimal values for the model parameters $\theta^*$ as well as the labels $\{y_j^*\}_{j=1}^{N}$ for the unlabeled examples $S_u$. Inspired by this, 
% Below is the underlying optimization problem formulation.
% \begin{eqnarray}
% & \theta^*, \{y_j^*\}_{j=1}^{N}  = \underset{\theta, \{y_j\}_{j=1}^{N} }{\arg\min} \;{\cal{L}}\left(S_{\ell}, S_u, \{y_j\}_{j=1}^{N}; \theta\right)\nonumber \\
% & = \underset{\theta, \{y_j\}_{j=1}^{N} }{\arg\min} \left[\sum\limits_{i \in S_{\ell}} \frac{C_{\ell}L({\mathbf{x}}_i, y_i; \theta)}{|S_{\ell}|} + \sum\limits_{j \in S_u} \frac{C_{u}L({\mathbf{x}}_j, y_j;\theta)}{|S_u|} \right] \label{tranductive_opt}
% \end{eqnarray}
% where, $C_{\ell}$ and $C_u$ are hyper-parameters, and $L(\cdot, \cdot)$ is an appropriately defined loss function for the chosen model. 
%For example in the case of \cite{Joachims:1999}, it was hinge loss.
% This optimization problem is combinatorial in nature and \cite{Joachims:1999} had proposed a local search method for solving this problem which alternates between the model parameters $\theta$ and the labels $\{y_j\}_{j=1}^{N}$.
%In what follows, we discuss our proposed approach to address the label scarcity issue while training DANN. 
we first reduce the given DA problem into a semi-supervised learning problem over the common domain (feature space $\mathbf{f}$) and subsequently improve the task classifier $G_y(\mathbf{f};\hat{\theta}_y)$ in a similar way. The net effect is that resulting classifier outperforms the classifier $G_y(G_f(\mathbf{x};{\hat{\theta}}_f);{\hat{\theta}}_y)$ obtained by training the DANN.  
%%%%%%%%%%%%%%%%%%%%%%%%%%%%%%%%%%%%%%%%%%%%%%%%%%%%%%%%%%%%%%%%%%%%%%%%%%%%%%%%%

%
%\input{TransDANN_arxiv.tex}
%
%			          TransDANN - The Proposed Approach
%%%%%%%%%%%%%%%%%%%%%%%%%%%%%%%%%%%%%%%%%%%%%%%%%%%%%%%%%%%%%%%%%%%%%%%%%%%%%%%%%
\section{{\em TransDANN} -- The Proposed Approach}
\begin{figure*}[ht]
%\hspace*{0.2cm}
\centering
{\includegraphics[width=.8\linewidth, height =.37\linewidth]{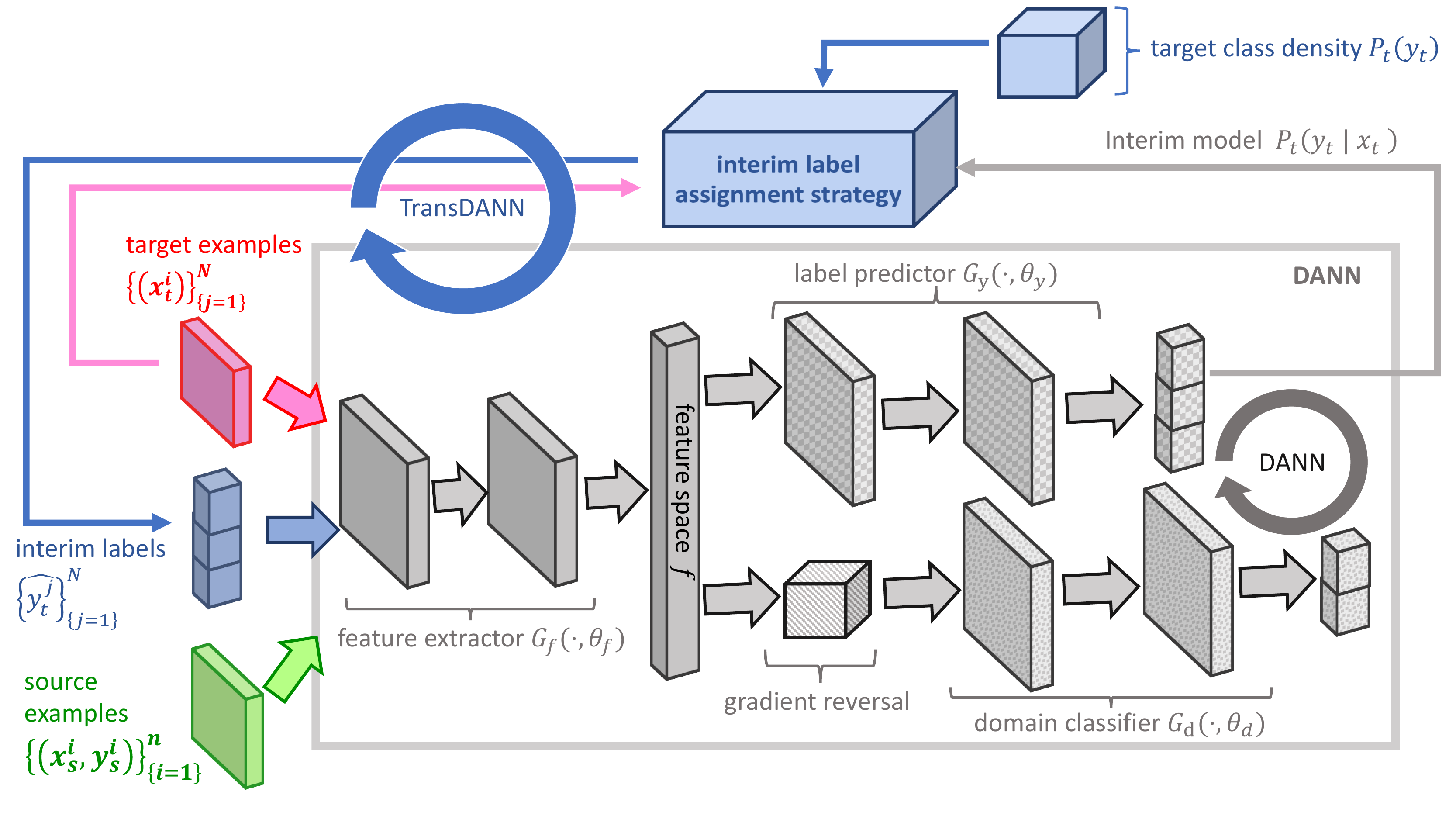}}
\vspace*{-0.35cm}
\caption{TransDANN Approach}
\label{fig:TransDANN}
\vspace*{-0.3cm}
\end{figure*}

In this section, we give details of our proposed modified training approach for DANN. We call this approach as {\em Transductive training of Deep Domain Neural Network (TransDANN)}. Figure \ref{fig:TransDANN} depicts the idea behind TransDANN approach.
%\begin{figure}[ht]
%%\hspace*{0.2cm}
%%\centering
%{\includegraphics[scale=0.26, angle=0]{TransDANN_Arch}}
%\vspace*{-0.3cm}
%\caption{TransDANN Approach}
%\label{fig:TransDANN}
%\vspace*{-0.3cm}
%\end{figure}

%Inspired by \cite{Joachims:1999}, 
In TransDANN approach, we being by defining the following alternative loss function
%for the purpose of training DANN under deep label scarcity. 
called as TransDANN loss function. ${\cal{L}}\left(\theta_f,\theta_y,\theta_d,D_s,D_t, \{y_t^j\}_{j=1}^{N}\right)=$
\begin{eqnarray}
%& {\cal{L}}\left(\theta_f,\theta_y,\theta_d,D_s,D_t, \{y_t^j\}_{j=1}^{N}\right)= \nonumber\\
& \sum\nolimits_{i=1}^{n} \frac{C_{\ell}L(h({\mathbf{x}}_s^{i};\theta_f, \theta_y), y_s^i)}{n}  + 
\sum\nolimits_{j=1}^{N} \frac{C_uL(h({\mathbf{x}}_t^{j};\theta_f, \theta_y), y_t^j)}{N}\nonumber \\
& - \lambda \left[\sum\nolimits_{i=1}^{n} \frac{L(h({\mathbf{x}}_s^{i};\theta_f, \theta_d), d_s^i=0)}{n} + \sum\nolimits_{j=1}^{N} \frac{L(h({\mathbf{x}}_t^{j};\theta_f, \theta_d), d_t^j=1)}{N}\right]\; \label{TransDANN_loss_func}     
\end{eqnarray}
where $L(\cdot,\cdot)$ is a cross-entropy loss function and $C_{\ell}, C_{u}$ are the importance weights. This loss function has flavor of both DANN loss function (given by Equation (\ref{loss_func})) and the transductive learning loss function (given in \cite{Joachims:1999}).
%the formulation (\ref{tranductive_opt}). 
The idea behind this loss function is to include the labels for the unlabeled examples (from the target domain) also as decision variables. As part of TransDANN training, we solve the following {\em saddle point problem}:
 %with respect to the TransDANN loss function defined above. That is, 
$\theta_f^*,\theta_y^* , \{y_t^{*^j}\}_{j=1}^{N} =$
\begin{eqnarray}
%& \theta_f^*,\theta_y^* , \{y_t^{*^j}\}_{j=1}^{N} = \nonumber \\ 
& \underset{\theta_f,\theta_y, \{y_t^{j}\}_{j=1}^{N}}{\arg\min} {\cal{L}}\left(\theta_f,\theta_y,\theta_d^*,D_s,D_t, \{y_t^j\}_{j=1}^{N}\right) \label{TransDANN_opt1}\\
& \theta_d^* = \underset{\theta_d}{\arg\max} \;\;{\cal{L}}\left(\theta_f^*,\theta_y^*,\theta_d,D_s,D_t, \{y_t^{*^j}\}_{j=1}^{N}\right) \label{TransDANN_opt2}
\end{eqnarray}
Observe, the first optimization problem (\ref{TransDANN_opt1}) is a combinatorial optimization problem. Therefore, unlike DANN, the overall saddle point problem also becomes a combinatorial optimization problem. 
%%%%%%%%%%%%%%%%%%%%%%%%%%%%%%%%%%%%%%%%%%%%%%%%%%%%%%%%%%%%%%%%%%%%%%%%%%%%%%%
%%% 						 Algorithm 1
%%%%%%%%%%%%%%%%%%%%%%%%%%%%%%%%%%%%%%%%%%%%%%%%%%%%%%%%%%%%%%%%%%%%%%%%%%%%%%%
\IncMargin{1em}
\begin{algorithm}[!htb]
\SetKwInOut{Input}{input}
\SetKwInOut{Output}{output}
\SetKwInOut{Initialization}{Initialize}
\Input{$D_s=\{({\mathbf{x}}_{s}^i, y_s^i)\}_{i=1}^n$, $D_t=\{{\mathbf{x}}_{t}^j\}_{j=1}^N$, $D_t^v=\{{\mathbf{x}}_{t}^j, y_t^j\}_{j=N+1}^{N+v}$, $\lambda^*, C_{\ell}^*, C_u^*$, $\{num_c\}_{1=1}^{k}$}
\Output{Model parameters $\theta_f^*, \theta_y^*, \theta_d^*$}
\BlankLine
\caption{Local Search Based Method for TransDANN Saddle Point Problem (\ref{TransDANN_opt1}) -- (\ref{TransDANN_opt2}) \label{TransDANN_Algo}}
Define a DANN with sufficiently rich feature map $G_f(\cdot, \theta_f)$ and domain classifier $G_d(\cdot,\theta_d)$\;
$C_{\ell} \leftarrow C_{\ell}^*,\; C_{u} \leftarrow 0, \; \lambda \leftarrow \lambda^*$ \;
Train the DANN on supplied data $D_s$ and $D_t$ \;
Let ${\theta}_f^{cold}, {\theta}_y^{cold}, {\theta}_d^{cold}$ be the model parameters of this trained DANN\;
%\tcc*[r]{Cold start complete}
$\hat{\theta}_f\leftarrow {\theta}_f^{cold}, \hat{\theta}_y\leftarrow {\theta}_y^{cold}, \hat{\theta}_d\leftarrow {\theta}_d^{cold}$\;
$C_{u} \leftarrow 10^{-3}$ \; 
%Invoke Algorithm \ref{Label_Shuffle_Algo} to get the labels $\{y_t^j\}_{j=1}^{N}$ for the target examples. Supply current label prediction model $h(\cdot \mid \mathbf{x})=G_y(G_f(\mathbf{x}; {\hat{\theta}}_f); {\hat{\theta}}_y)$ and $\{num_c\}_{1=1}^{k}$ while invoking Algorithm \ref{Label_Shuffle_Algo} \label{step_label_assignment}\;
%Sort all the target domain examples $D_t=\{{\mathbf{x}}_t^j\}_{j=1}^{N}$ in decreasing order of the score $G_y(G_f({\mathbf{x}}_t^j; {\hat{\theta}}_f);{\hat{\theta}}_y)$ \label{step_label_assignment}\;
%Assign a class label $y_t^j=1$ to top $num_+$ examples and a class label of $y_t^j=0$ to the remaining examples\;
\Repeat{$C_u < C_u^*$ \label{reapet_step}} 
{
	Invoke Algorithm \ref{Label_Shuffle_Algo} on $G_y(G_f(\mathbf{x}; {\hat{\theta}}_f); {\hat{\theta}}_y)$ and $\{num_c\}_{1=1}^{k}$ to get interim labels $\{{\hat{y}}_t^j\}_{j=1}^{N}$ \label{label_revision_step}\;
	Start with current values of ${\theta}_f, {\theta}_y, {\theta}_d$ and retrain DANN by including interim labels of  target examples, namely $\{(\mathbf{x}_t^j, {\hat{y}}_t^j)\}_{j=1}^{N}$ \label{parameter_revision_step}\;
	Let $\hat{\theta}_f, \hat{\theta}_y, \hat{\theta}_d$ be the revised values for the DANN parameters after this training\;
	$C_u \leftarrow \min\{2*C_u, C_u^*\}$
	%Go to Step  \ref{step_label_assignment}
}
Evaluate the model $G_f(G_y(\mathbf{x}; \theta_y^{cold});\theta_f^{cold})$ as well as 
$G_f(G_y(\mathbf{x}; \hat{\theta}_y);\hat{\theta}_f)$
on validation set $D_t^v$\;
Whichever model performs better, output the corresponding parameters as $\theta_f^*, \theta_y^*, \theta_d^*$ 
\end{algorithm}
%%%%%%%%%%%%%%%%%%%%%%%%%%%%%%%%%%%%%%%%%%%%%%%%%%%%%%%%%%%%%%%%%%%%%%%%%%%%%%%
%%% 							End of Algorithm 1
%%%%%%%%%%%%%%%%%%%%%%%%%%%%%%%%%%%%%%%%%%%%%%%%%%%%%%%%%%%%%%%%%%%%%%%%%%%%%%%
Our proposed method to solve this saddle point problem is presented in the form of Algorithm \ref{TransDANN_Algo} 
% \begin{wrapfigure}[20]{I}{1.6\columnwidth}
% \includegraphics[width=1.6\columnwidth]{TransDANN_Arch}
% \end{wrapfigure}
%At a high level, the method given in  Algorithm \ref{TransDANN_Algo} 
and it works as follows. It starts with a small number of labeled examples $D_s=\{({\mathbf{x}}_{s}^i, y_s^i)\}_{i=1}^n$ from the source domain and and a large number of unlabeled examples $D_t=\{{\mathbf{x}}_{t}^j\}_{j=1}^N$ from the target domain. As a cold start, the method temporarily ignores the variables $\{y_t^{j}\}_{j=1}^{N}$ (and hence the second term in the Equation (\ref{TransDANN_loss_func})). Instead, it trains a vanilla DANN 
%(keeping sufficiently large capacity for the feature map and the domain classifier) 
on the given data so as to acquire an initial assignment of the $\theta$-parameters, given by ${\theta}_f^{cold},{\theta}_y^{cold},{\theta}_d^{cold}$. Next, there is a loop which alternates between variables $(\theta_f, \theta_y, \theta_d)$ and $\{y_t^{j}\}_{j=1}^{N}$ so as to improve them in lieu of the sub-problems (\ref{TransDANN_opt1}) -- (\ref{TransDANN_opt2}). That means, in one step (Step \ref{parameter_revision_step}), it clamps the current assignment of the labels $\{y_t^{j}\}_{j=1}^{N}$ and improves upon the parameters $(\theta_f,\theta_y,\theta_d)$ in their local vicinity. In the subsequent step (Step \ref{label_revision_step}), it clamps $(\theta_f,\theta_y,\theta_d)$ to their present values and revises the labels $\{y_t^{j}\}_{j=1}^{N}$ so as to reduce the loss. We call these revised labels as {\em interim labels} and denote them by $\{{\hat{y}}_t^{j}\}_{j=1}^{N}$. 

In this alternation strategy, when improving upon the parameters $(\theta_f,\theta_y,\theta_d)$ locally, we follow a DANN like strategy
%alternation between problems (\ref{TransDANN_opt1}) and (\ref{TransDANN_opt2}) 
because the second term of the TransDANN loss function is a constant. On the other hand, when improving upon the parameters $\{y_t^{j}\}_{j=1}^{N}$, we need to solve the sub-problem (\ref{TransDANN_opt1}) clamping the variables $({\theta}_f,{\theta}_y,{\theta}_d)$ to their current values. Because this sub-problem (\ref{TransDANN_opt1}) is a combinatorial optimization problem (due to the presence of $\{y_t^j\}$), we advocate the use of a local search strategy for this sub-problem. By local search strategy, we mean that we greedily revise the current assignment of the labels for $\{y_t^{j}\}_{j=1}^{N}$ so as to reduce the overall value of the loss function (\ref{TransDANN_loss_func}). In the next section, we describe one such strategy to assign interim labels.

In Algorithm \ref{TransDANN_Algo}, as iterations proceed, we slowly increase the importance weight $C_u$ 
%for the target domain classification loss 
until it hits the user specified upper bound $C_u^*$. The value of $C_u^*$ dictates how much importance we wish to give to the semi-supervised part. Finally, suppose Algorithm \ref{TransDANN_Algo} is given access to a validation set $D_t^v=\{{\mathbf{x}}_{t}^j, y_t^j\}_{j=N+1}^{N+v}$ -- a small labeled set from the target domain. In this situation, it compares the performance of the cold start model $\theta_f^{cold}, \theta_y^{cold}, \theta_d^{cold}$ (offered by vanilla DANN) with the TransDANN model $\hat{\theta}_f, \hat{\theta}_y, \hat{\theta}_d$ obtained at the end of iterative loop 7--12. The algorithm outputs a better of these two models, denoted by $\theta_f^*, \theta_y^*, \theta_d^*$.      
\subsection{Interim Label Assignment Strategy}
As far as the revision of the labels $\{y_t^{j}\}_{j=1}^{N}$ is concerned in Algorithm \ref{TransDANN_Algo}, there could be many strategies but we opt the following strategy which we call as {\em matching the class distribution} strategy. The idea behind this strategy is to assign the labels $\{y_t^{j}\}_{j=1}^{N}$ to the target domain examples in a way that these labels are in sync with the current label prediction model, call it  $h(y\mid\mathbf{x})=P(y\mid\mathbf{x})$, {\em as much as possible}, and at the same time, the distribution of the labels across the classes adhere to some apriori given numbers $\{n_c\}_{c=1}^{k}$, where $n_c = N \times {P}_t(y=c)$. The class densities ${P}_t(y)$ are assumed to be either known or equal to ${P}_s(y)$ which can be estimated from the source labeled examples. The reason being that throughout the TransDANN, induced marginals in the feature space remain the same and the label predictor $G_y(\cdot)$ also remains the common between source and target domains. Therefore, ${P}_s(y)={P}_t(y)$ all the times.  

For the general scenario of $k \ge 2$, this strategy is given in the form of Algorithm \ref{Label_Shuffle_Algo}. This algorithm works as follows. First, we assign each example to the best class as per the supplied label prediction model $P(y\mid\mathbf{x})$. Next, we pick some class $c$ which has the surplus number of examples relative to its target $n_c$. Among all the examples assigned to this class $c$, we identify the one which has the weakest membership score $P(y=c\mid\mathbf{x})$ and move that example to some other class, say $\hat{c}$. The class $\hat{c}$ is chosen such that it has a deficiency of examples relative to its target $n_{\hat{c}}$ and moreover, the identified example has strongest membership score for this class as compared to other classes who also have a deficiency.      
%For the case of $k=2$, this strategy becomes quite simple and reads like this. Sort all the unlabeled examples as per the current model's score $G_y(G_f(\mathbf{x}_t^j;\theta_f);\theta_y)$. These scores correspond to the $P(y=1|\mathbf{x})$. From this sorted list, pick the top $n_1$ examples and assign their revised labels to be 1. The rest of the examples are assigned $0$ as their revised label. 
%The quantity $n_+$ is obtained from the prior information about the class distribution, that is $num_+ = N \times \mathbb{P}_t(y=1)$. If this class distribution is not known upfront then we estimate it from the labeled examples because it should be the same as $\mathbb{P}_s(y=1)$. 
%%%%%%%%%%%%%%%%%%%%%%%%%%%%%%%%%%%%%%%%%%%%%%%%%%%%%%%%%%%%%%%%%%%%%%%%%%%%%%%
%%% 							Algorithm 2
%%%%%%%%%%%%%%%%%%%%%%%%%%%%%%%%%%%%%%%%%%%%%%%%%%%%%%%%%%%%%%%%%%%%%%%%%%%%%%%
\IncMargin{1em}
\begin{algorithm}[!htb]
\SetKwInOut{Input}{input}
\SetKwInOut{Output}{output}
\SetKwInOut{Initialization}{Initialize}
\Input{ $\{\mathbf{x}_t^j\}_{j=1}^{N}$ whose labels needs revision,
$P(y \mid \mathbf{x}); \forall \mathbf{x} \in {\cal{X}}, y=1\rightarrow,k$, Class distribution $\{n_c\}_{c=1}^{k}$ 
%or class density $\{{P}_t(y=c)\}_{c=1}^{k}$.
}
\Output{Labels $\{{\hat{y}}_t^j\}_{j=1}^{N}$.}
\BlankLine
\caption{Interim Label Assignment \label{Label_Shuffle_Algo}}
% \If{target densities are given}
% {
% 	$n_c \leftarrow N \times P_t(y=c)\;\forall c=1 \rightarrow k$
% }
$S_c \leftarrow \emptyset \;\forall c=1\rightarrow k$\;
%\tcc*[r]{For assigning examples to the classes}
$U^+ \leftarrow \emptyset, \; U^- \leftarrow \emptyset$ \tcc*[r]{For tracking unbalanced classes}
\For{$j=1 \rightarrow N$}
{
	$c_j = \underset{c=1\rightarrow k}{\arg\max}\;\;P(y=c \mid{\mathbf{x}}_t^j)$\;
	$S_{c_j} \leftarrow S_{c_j} \cup \{{\mathbf{x}}_t^j\}$ 
}
\For{$c=1 \rightarrow k$}
{
	$U^+ \leftarrow U^+ \cup \{c\}\;\text{if} \; |S_c| > n_c\;$\;
	$U^- \leftarrow U^- \cup \{c\}\;\text{if} \; |S_c| < n_c$\;
	% \If{$|S_c| > n_c$}
	% {
	% 	$U^+ \leftarrow U^+ \cup \{c\}$
	% }
	% \If{$|S_c| < n_c$}
	% {
	% 	$U^- \leftarrow U^- \cup \{c\}$
	% }
}

\Repeat{$|U^+| >0 $}
{
	Let $c$ be some element of $U^+$\;
	${\mathbf{x}}_t^* \leftarrow \underset{{\mathbf{x}}_t\in S_c}{\arg\min}P(y=\hat{c}\mid {\mathbf{x}}_t)$ \;
	%Let ${\mathbf{x}}_t^* \in S_c$ be an element for whom the score $P(y=c\mid {\mathbf{x}}_t^*)$ is the minimum\;
	Find a class $\hat{c}\in U^-$ for which the score $P(y=\hat{c}\mid {\mathbf{x}}_t^*)$ is the maximum\;
	$S_c \leftarrow S_c \setminus \{{\mathbf{x}}_t^*\}$;
	$S_{\hat{c}} \leftarrow S_{\hat{c}} \cup \{{\mathbf{x}}_t^*\}$\;
	$U^+  \leftarrow U^+ \setminus \{c\} \;\text{if} \;|S_c| \le n_c$\;
	$U^-  \leftarrow U^- \setminus \{\hat{c}\} \;\text{if} \;|S_{\hat{c}}| \ge n_{\hat{c}}$\;
	% \If{$|S_c| \le n_c$}
	% {
	% 	$U^+  \leftarrow U^+ \setminus \{c\}$
	% }
	% \If{$|S_{\hat{c}}| \ge n_{\hat{c}}$}
	% {
	% 	$U^-  \leftarrow U^- \setminus \{\hat{c}\}$
	% }   
}
\For{$j=1\rightarrow N$}
{
	${\hat{y}}_t^j \leftarrow c_j \;\text{if} \;{\mathbf{x}}_t^j \in S_{c_j}$
}
Output $\{{\hat{y}}_t^j\}_{j=1}^{N}$
\end{algorithm}
\subsection{Theoretical Analysis of TransDANN}
%In this section, we state a key result stating performance guarantee for the proposed TransDANN approach. 
Theorem given below guarantees  
%This theorem guarantees 
that under some mild conditions, model learned by the TransDANN is no inferior than DANN.
 %if not better under most of the scenarios. 
 %This is verified in our experiments as well. 
 The proof %(given in supplementary material) 
 relies on a fact that DANN ignores the term $\lambda$ while minimizing the error bound of Theorem \ref{ben-david-theorem}.  
\begin{theorem} Suppose covariance shift assumption holds true and one can solve the DANN saddle point problem given in (\ref{saddle_eqn_1}) -- (\ref{saddle_eqn_2}), then it's unlikely that the TransDANN algorithm \ref{TransDANN_Algo} would learn a model $P(y\mid \mathbf{x})$ that is inferior to the model learned by DANN for the same input dataset.
\end{theorem}

{\bf Proof : }
Recall, DANN essentially tries to minimize the error bound given in Theorem 2. However, it tries to minimize the sum of only first two terms in the error bound of Theorem 2 and 
ignores the last term $\lambda$ by treating it as a constant. We would like to highlight that the term $\lambda$ is defined as $\underset{h \in {\cal{H}}}{\min}\left(\epsilon_s(h)+\epsilon_t(h)\right)$. In the case of DANN, the hypothesis space $\cal{H}$ is controlled by both $\theta_f$ and $\theta_y$ parameters. However, in DANN's training, the update  of $\theta_f$ and $\theta_d$ is never influenced by $\lambda$. The reason behind this is also apparent  -- DANN assumes no supply of the labeled data from the target domain and hence it has no way to estimate $\lambda$ with reasonable accuracy. 

On the other hand, in TransDANN, we indirectly estimate the term $\lambda$ by the inclusion of a term capturing the label classification loss in the target domain. This term, in conjunction with the label classification loss for the source domain, mimics $\lambda$. In order to calculate this term, we need labels for the target domain which we get from the interim label assignment layer in the TransDANN. In the initial iterations, these interim labels are not accurate and hence the estimation of $\lambda$ is poor. However, as iterations progress, the interim labels for the target domain examples improves and so does the estimation of $\lambda$. This helps TransDANN get an improved lower error bound than what DANN would get. Also, for the above argument to hold, we need covariate shift assumption because in each iteration of the TransDANN, we estimate interim labels of the target domain by using the current model for the source label. If covariate shift assumption is not true then we can't assume that the labels estimated by interim label assignment layer during TransDANN training would eventually be trustworthy to get a good estimate of $\lambda$.
%%%%%%%%%%%%%%%%%%%%%%%%%%%%%%%%%%%%%%%%%%%%%%%%%%%%%%%%%%%%%%%%%%%%%%%%%%%%%%%          

%
%\input{exp.tex}
%
%%% 						Experiments and Results
%%%%%%%%%%%%%%%%%%%%%%%%%%%%%%%%%%%%%%%%%%%%%%%%%%%%%%%%%%%%%%%%%%%%%%%%%%%%%%%          

\section{Experimental setup}
%
% As discussed earlier, our approach can be applied to any DA task so long as the marginal distributions of the source and the target domains when induced in a latent space are reasonably close \cite{GaninUAGLLML16,Tzeng2017}. Because of this, we consider DANN as the base architecture and build the TranDANN approach on top of it as shown in Figure \ref{fig:TransDANN}.  Moreover, the proposed TransDANN approach  is applicable for domain adaptation irrespective of image or text data. 
To conduct an extensive set of experiments across various domains, we choose  Amazon review dataset -- a popular dataset among multi-domain deep learning (DL) methods \cite{huang2015,multinomial,GaninUAGLLML16}. We also experiment with MNIST and MNIST-M \cite{GaninUAGLLML16} datasets
%, and OFFICE datasets \cite{Senko2010}, 
which are commonly used for DA tasks in computer vision. %(but have fewer  images). 
%These are considered a { \em de facto} standard datasets for DL models in text and images.

\subsection{Dataset}

For DA on text experiments, we work with Amazon review dataset~\footnote{https://www.cs.jhu.edu/?mdredze/ datasets/sentiment/}. This dataset comprises of customer reviews (in the text form) for 14 different product-lines (aka domains) at Amazon including Books, DVDs, Music, etc. The labels correspond to the sentiments of the reviewers. We extracted sentences and their corresponding labels from the raw data provided by \cite{Blitzer07}. We processed the sentences using Stanford tokenizer \footnote{http://nlp.stanford.edu/software/ tokenizer.shtml}.   
For each domain, the data are partitioned randomly into {\em training, development,} and {\em test} sets in a ratio of 70\%, 10\%, 20\%, respectively. 
%Table \ref{t:dataset} presents a summary statistics about this dataset. 
For our experimentation, we selected only 10 of these domains and hence we have skipped the details of the 4 domains from this tables as well as subsequent results. The detailed statistics of this dataset is given in supplementary material.
For DA on images, we experiment with MNIST dataset available \cite{Lecun1998} as source and MNIST-M, obtained from  \cite{GaninUAGLLML16}, as target domains. 
\subsection{Baselines}
Our proposed approach aims to improve the {DANN} performance for DA tasks. Therefore, we treat the performance of the {DANN} as a baseline for our experiments.  In our experiments, for each source-target domain pair, a baseline DANN model is trained as suggested in \cite{GaninUAGLLML16}. In addition, to get an idea of how good the DANN itself perform in the first place, we also train a {\em target-only} model. We train such a target-only model using only the task classifer part of DANN architecture with  labeled examples only from the target domain. %The performance of such a model serves as a baseline for the performance of any deep net based approach.
%even when there is a considerable shift between source and target domains. %[We display in diagonal terms]. 

To emulate the label scarcity, we restrict the supply of labeled data from the source domain.  
% we limit the revealed source data labels for training each source-target pair, called as \textbf{Label Scarce} (LS) scenario. 
Under such label scarcity (LS) scenarios, the performance of DANN deteriorates as shown in Figure \ref{fig:DANN_Detoriation}. 
%As mentioned earlier, performance of DANN model in LS scenarios deteriorates, illustrated in Fig. \ref{fig:DANN_Detoriation}.  
Our proposed approach, TransDANN, aims to achieve better performance (target accuracy) than DANN, especially under the LS scenarios.  For relative comparison, we train the models using both DANN as well as TransDANN approaches under different LS scenarios.
%by limiting the supply of source labeled examples at different levels. 
%We train TransDANN in LS scenario with different levels of source labels scarcity and show the performance improvement over DANN model.
%we evaluate our proposed algorithm for each source-target pair under varying levels of source labels supply (ranging from 100\% to 80\%)
For text domain experiments,
%In our experiments, 
we limited the supply of the source labeled data ranging from 100\% to 80\% in each of source-target domain pair. 
Similarly, the image experiments are carried over the number of examples ranging 10000 to 4000 in each of source-target pair, i.e., MNIST and MNIST-M dataset.  We found that performance of DANN remains similar with any number of examples more than 10000. For MNIST, we keep target supply same as that of source. We found that these range of label examples are required for reasonably good deep feature extraction. 
%We chose this range of label scarcity because we found that the available labels to train target-only model are apt for deep feature extraction when we stick to this range of LS. 
%Note, if source data labels are few, the DANN like models would seize to train even without domain adapter branch. % 
%
%
%%%%%%%%%
\setlength{\tabcolsep}{0.25em}
\begin{table*}[t]
\footnotesize
%\centering
\parbox{0.5\textwidth}{
%\begin{adjustbox}{angle=90}
\begin{tabular}{lcccccccccc}
\toprule 
{Target} &  dvd &  books &  elect &  baby &  kit &  music &  sports &  app &  cam &  health \\

{Source} &   &&   &   &  &  & &   &   &  \\
\midrule
\rowcolor{Gray}
dvd    &      &  75.0 &  74.2 & 77.0 & 75.8 &  79.1 &   62.5 & 82.6 & 71.5 &   55.3 \\
books  & 83.0 &       &  78.9 & 74.6 & 78.9 &  80.3 &   79.7 & 81.4 & 81.4 &   72.9 \\
\rowcolor{Gray}
elect  & 74.4 &  74.2 &       & 79.1 & 79.3 &  73.4 &   81.6 & 82.0 & 74.6 &   77.1 \\
baby   & 63.9 &  71.3 &  80.9 &      & 82.4 &  70.3 &   79.1 & 79.5 & 77.0 &   54.5 \\
\rowcolor{Gray}
kit    & 70.7 &  69.3 &  77.1 & 77.0 &      &  63.9 &   80.5 & 82.2 & 75.8 &   80.1 \\
music  & 80.5 &  78.9 &  75.4 & 74.2 & 74.8 &       &   79.5 & 80.3 & 79.9 &   74.2 \\
\rowcolor{Gray}
sports & 72.1 &  69.7 &  80.7 & 82.2 & 85.4 &  74.2 &        & 85.4 & 82.2 &   81.4 \\
app    & 72.9 &  69.9 &  79.5 & 80.5 & 79.5 &  72.9 &   73.6 &      & 76.4 &   78.5 \\
\rowcolor{Gray}
cam    & 76.8 &  71.3 &  80.3 & 74.4 & 77.7 &  71.5 &   80.1 & 78.1 &      &   80.9 \\
health & 70.3 &  71.9 &  82.2 & 83.0 & 81.4 &  77.3 &   80.7 & 84.6 & 80.8 &        \\
\midrule
T-O &	83.9	&88.0  & 81.4 & 83.9 & 81.8   & 78.6   &   85.1    & 80.6    & 83.7    & 84.5\\

\bottomrule
\end{tabular}
%\end{adjustbox}
\caption{ Baseline target test accuracy of DANN trained on various source-target domains on Amazon review dataset. T-O (target-only) model is trained on target true label revealed on the same dataset.}
\label{t:res_dann}
}
\qquad\qquad
\begin{minipage}[c]{0.43\textwidth}%
%\centering
    \includegraphics[width=1\textwidth]{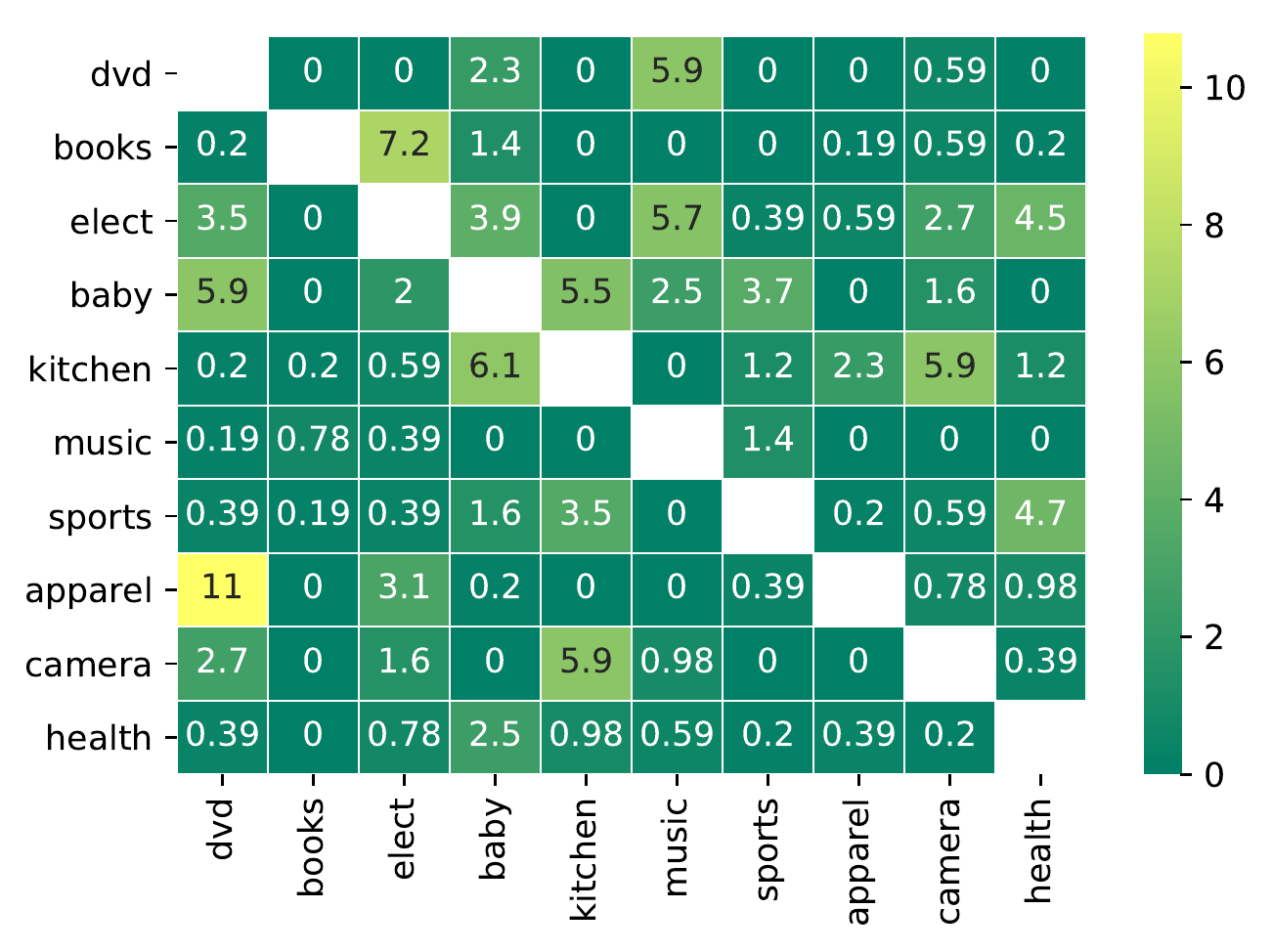}
\caption{Max \% improvement in accuracy of TransDANN over DANN (y-axis $\rightarrow$ source and x-axis $\rightarrow$ target)}
\label{fig:acc_imp}
\end{minipage}
\vspace*{-0.35cm}
\end{table*}
%%%%%%%%%%%%%
\subsection{Architecture}
The proposed TransDANN approach comprises two main pieces --  (i) DANN, and (ii) Interim Label Assignment. 

DANN consists of feature extractor, task classifier, and domain adaptation layer.  
Feature extractor for the {\em text domain adaptation} can be composed of neural sentence models such as recurrent neural networks \cite{sutskever2014sequence,Chung2014,Liu2015}, convolution networks~\cite{Collobert2011,Kalchbrenner2014}, or recursive neural networks \cite{Socher2013}. Here, we adopt recurrent neural network with long short-term memory (LSTM) due to their superior performance in various NLP tasks~\cite{Liu2016,Lin2017}. Specifically, we compose feature extractor with a bidirectional-LSTM and task classifier with a fully connected layer -- both as per the configurations suggested in the previous work on text modeling~\cite{Liu2017,Lin2017}.  Preprocessing and tokenization of the input sentences are carried out as suggested by the standard NLP text modeling methods \cite{Liu2017}. The words embedding for all the models are initialized with the 300-dimensional GloVe vectors~\cite{Pennington2014}. Other parameters are initialized by randomly sampling from a uniform distribution in the range $[-0.1,0.1]$. 
For the domain adapter component, we stick to three fully connected layers ($x\rightarrow 1024 \rightarrow 1024 \rightarrow 2$) as suggested by \cite{GaninUAGLLML16}. 
For the image experiment, we use small CNN architecture for feature extractor, and 2 layer domain adapter ($x\rightarrow 100 \rightarrow 2$) exactly as in \cite{GaninUAGLLML16} .
We choose cross-entropy and logistic regression loss for task classification and domain adapter, respectively. 

The Interim Label Assignment layer assigns the target labels based on Algorithm \ref{Label_Shuffle_Algo}, which are then fed to the input. 

\subsubsection{Training Procedure}

The training under TransDANN proceeds in cycles. In each cycle, all the training examples are used in batches. 
The first cycle is purely DANN training and the interim labels for target examples kick in from the second cycle on-wards. Since true target labels are not available, the first cycle is trained simply as vanilla DANN wherein, the input batches are composed of labeled source examples and unlabeled target examples. From the second cycle onwards, the input to the model consists of interim target labels in addition to the source labels along with source and target examples. The interim target labels are generated by Interim Target Label Assignment (ITA) layer in the beginning of each cycle (except for the first cycle). The ITA layer ingests the trained model of the previous cycle and target class distribution so as to compute the new interim target labels using Algorithm \ref{Label_Shuffle_Algo}. These iterative cycles continue till the convergence of target label training accuracy. 

The model is trained on 128 sized batches for text and 64 sized for image. 
%For text data training, texts are preprocessed, tokenized in standard manner before training. 
Half of each batch is composed of samples from the source domain and the other half with samples from the target domain. In the very first cycle, where we train vanilla DANN, we increase the domain adaptation factor $\lambda$ slowly during early stage of the training so as to suppress noisy labels inferred by the domain classifier. 
\subsection{Choosing meta-parameters} 
The TransDANN training requires choosing meta-parameters ($\lambda$, learning rate, momentum, network architecture) in an unsupervised manner. One can assess the performance of the whole system (including the effect of hyper-parameters) by observing the error on a held-out data from the source domain as well as the training error on domain classification. For most of the meta-parameters, we have followed the guidelines from \cite{GaninUAGLLML16}. In general, we have observed good correspondence between the DA task performance and the performance on the held-out data from the source domain  which is in congruence with \cite{GaninUAGLLML16}. 
\section{Evaluation Results and Analysis}
%
%We now discuss various results and their analysis.
%For our dataset, we first evaluate the performance
%%%%%%%%%%%%%%%
% \begin{figure}[h]
% %\hspace*{0.2cm}
% \centering
% {\includegraphics[scale=0.5]{images/Improvement_heatmap}}
% \vspace*{-0.3cm}
% \caption{Percentage accuracy improvement of TransDANN over DANN (y-axis $\rightarrow$ source and x-axis $\rightarrow$ target)}
% \label{fig:acc_imp}
% \vspace*{-0.3cm}
% \end{figure}
%%%%%%%
%%%%%%
%\begin{figure}[h]
%%\hspace*{0.2cm}
%\centering
%{\includegraphics[scale=0.5]{images/Transdann_1}}
%%\vspace*{-0.3cm}
%\caption{Comparison of TransDANN vs DANN over held-out labeled set from target domain for varying amount of source labels}
%\label{fig:TansDANN_vs_DANN_for_individual_pairs}
%%\vspace*{-0.5cm}
%\end{figure}
%%%%
\begin{figure}
\centering
        \begin{subfigure}[b]{0.23\textwidth} 
        \includegraphics[width=\linewidth, height =.55\linewidth]{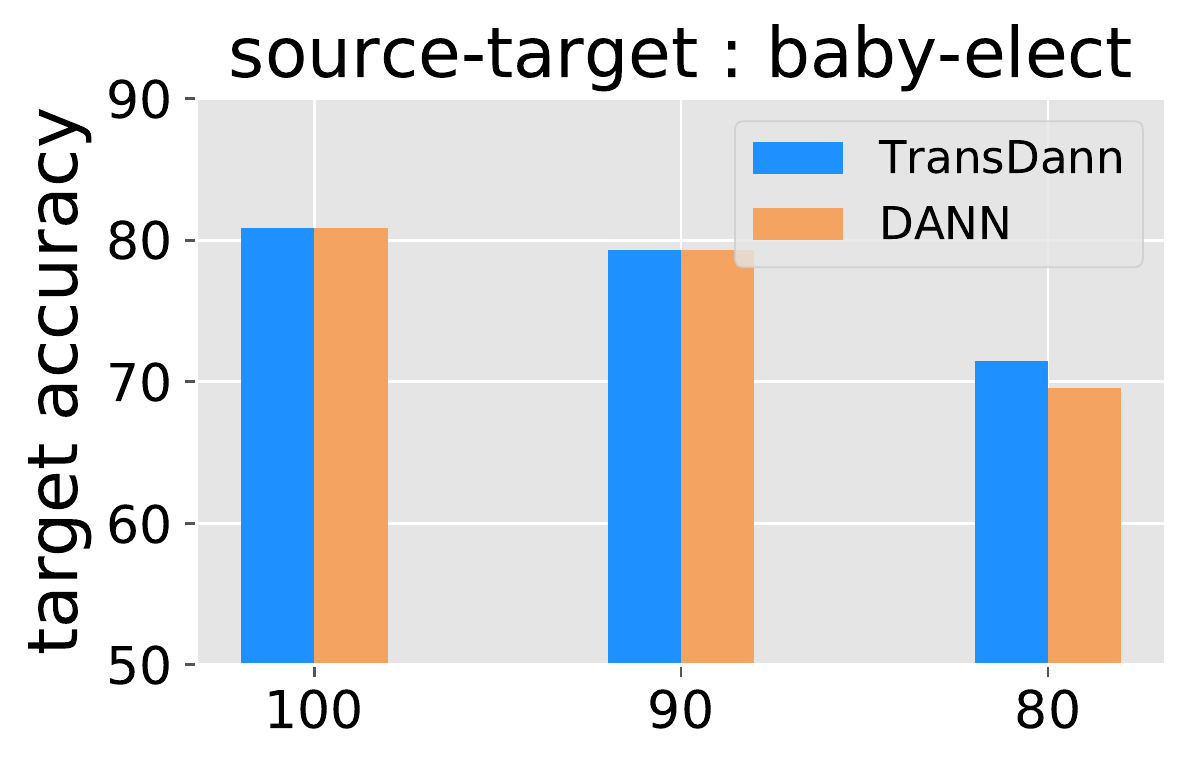}\end{subfigure}%
        \hspace{\fill}
        \begin{subfigure}[b]{0.23\textwidth}
                \includegraphics[width=\linewidth, height =.55\linewidth]{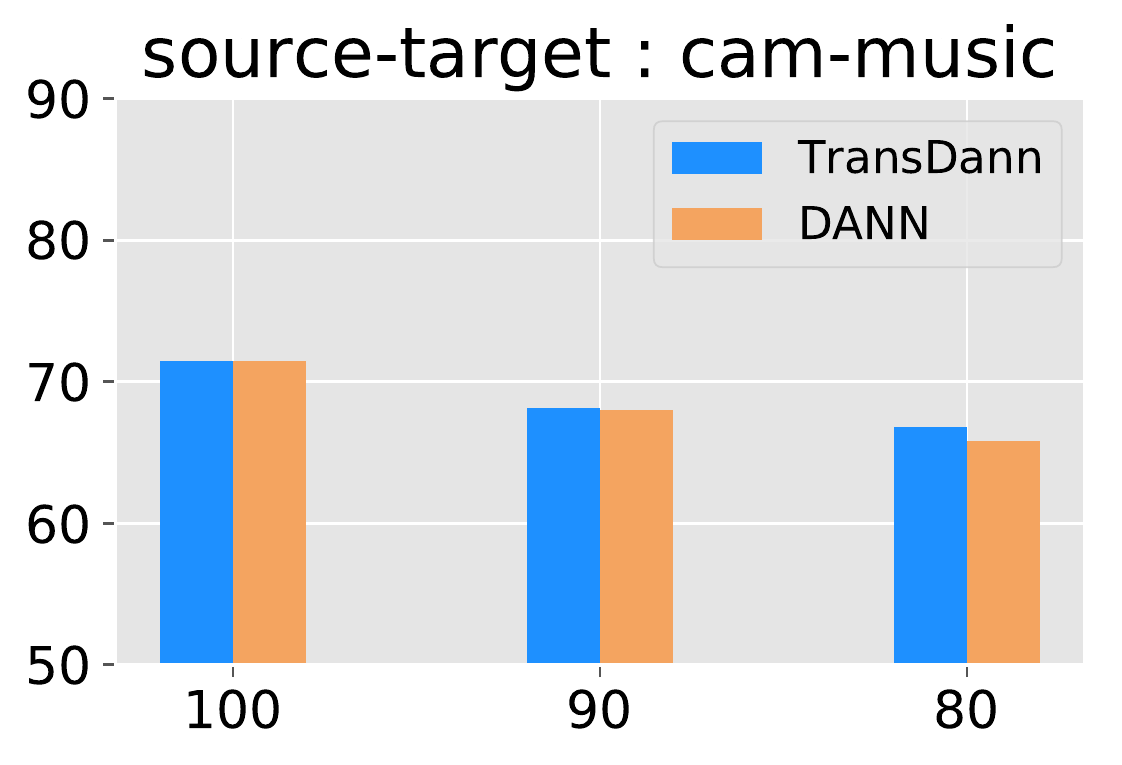}
         \end{subfigure}%
        \hspace{\fill}
        \begin{subfigure}[b]{0.23\textwidth}
                \includegraphics[width=\linewidth, height =.55\linewidth]{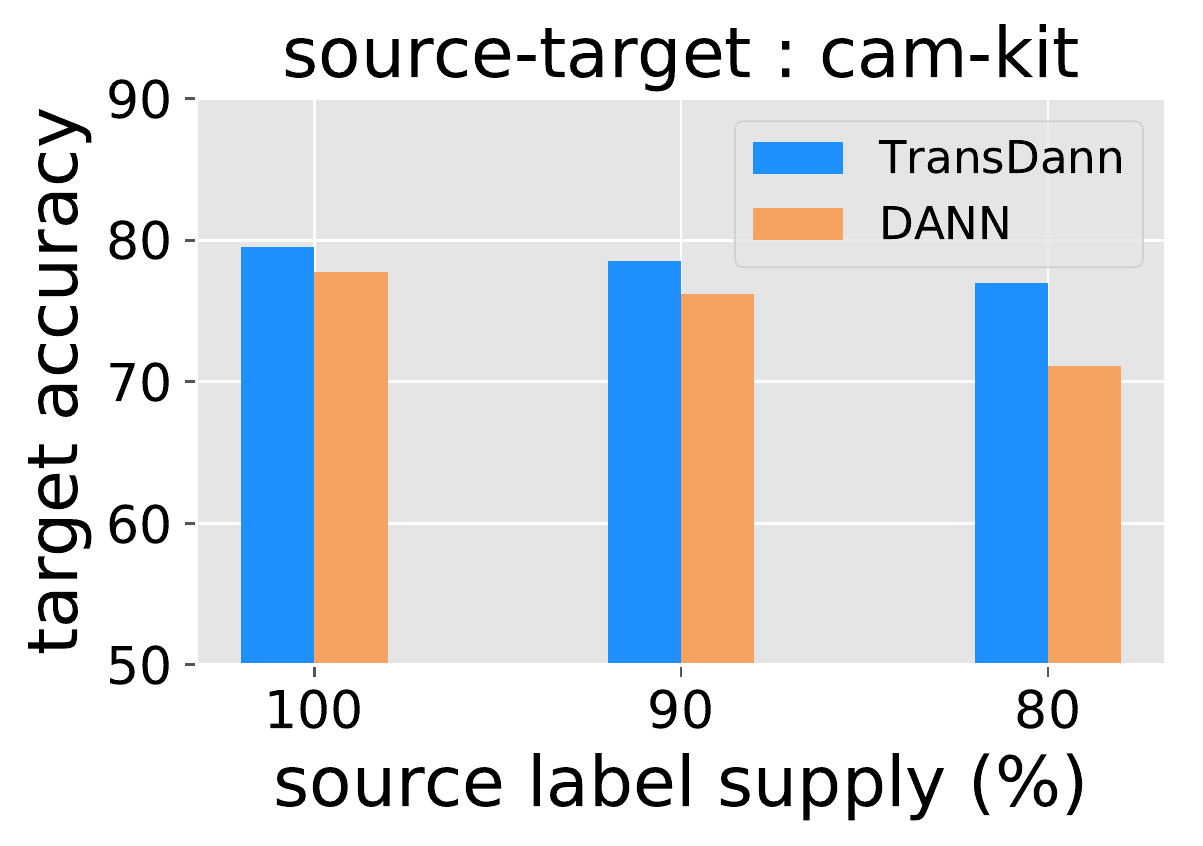}
        \end{subfigure}%
        \hspace{\fill}
        \begin{subfigure}[b]{0.23\textwidth}
                \includegraphics[width=\linewidth, height =.55\linewidth]{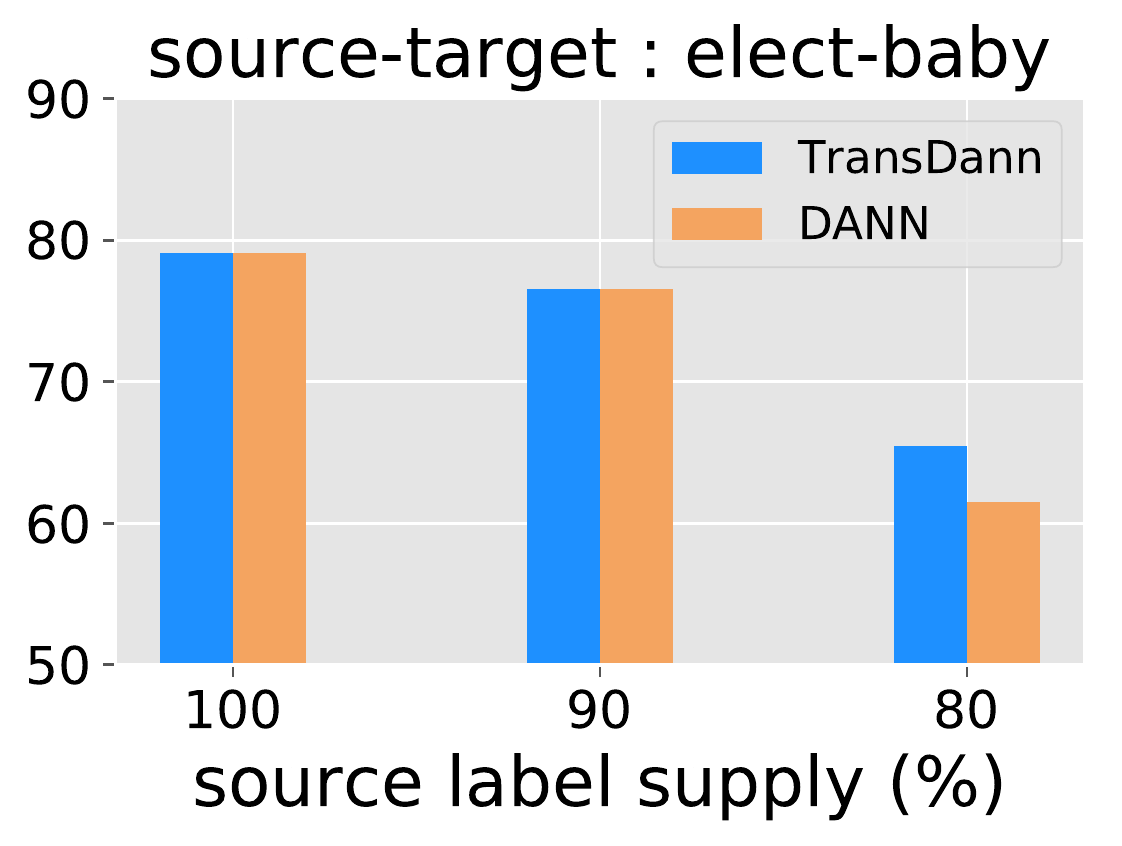}
        \end{subfigure}
        \vspace*{-0.15cm}
        \caption{Comparison of TransDANN vs DANN over held-out labeled set from target domain for varying amount of source labels}
        \label{fig:TansDANN_vs_DANN_for_individual_pairs}
\vspace*{-0.35cm}
\end{figure}
%
%%%%
%We first describe the performance of our algorithm on text data.  
{ \bf On Review dataset:} We describe the evaluation results on text data in the following. 
%Having established the baseline performance, 
We first obtain the baseline and then evaluate performance of {\em TransDANN } %our proposed algorithm 
for each source-target pair under varying levels of source labels supply (ranging from 100\% to 80\%). Table \ref{fig:acc_imp} summarizes the maximum \% improvement (over varying levels of source labels supply) achieved by TransDANN over DANN. Note, the diagonal elements are blank because experiments are conducted only for source-target pairs where $source \not= target$ so as to capture the efficacy of the proposed approach for DA tasks. The supplementary material contains the actual accuracy numbers over which these maximum \% improvements are calculated.   

A few important observations can be made from the Table \ref{fig:acc_imp}. First, it's clear that TransDANN outperforms DANN in several cases ($>70$\% cases) by a noticeable margin. Second, in cases where TransDANN performance is close or equal to DANN, we often found that the performance of the DANN on {\em target-only} task is either too bad or too good. When performance of the DANN itself is too bad on the {\em target-only} task, there could potentially be issues other than the label scarcity, for example, {\em covariate shift} may not be holding true. In such cases, we anyways can't expect TransDANN to improve significantly over DANN. On the other hand, when performance of the DANN itself is too good on the {\em target-only} task, there is not much scope for the TransDANN to improve.      
%a few insights. Firstly, TransDANN helps to improve the performance deficiency of DANN generally when it is caused due to scarcity of source labels. The cases where DANN also performs low at full source label supply, performance of TransDANN is also not appreciable, which can attributed to covariate shift assumptions doesn't hold true firmly with these domains pairs. On the other hand, DANN achieves good accuracy while TransDANN shows hardly any improvement over DANN, which is mainly because these source-target pair domains have small shift and covariate shift is held firmly. 

Overall, the accuracy improvement of TransDANN over DANN is found to be significant under the scenarios  where performance of the DANN gets affected due to the reduced supply of the source labels while the assumption of covariate shift holds true. To support this argument, we depict the performance of TransDANN over DANN with varying levels of source label supply in Figure \ref{fig:TansDANN_vs_DANN_for_individual_pairs} (refer supplementary material for enlarged version).

\begin{figure}
\vspace*{-0.15cm}
\centering
        \begin{subfigure}[b]{0.4\textwidth} 
        \includegraphics[width=0.9\linewidth, height =.5\linewidth]{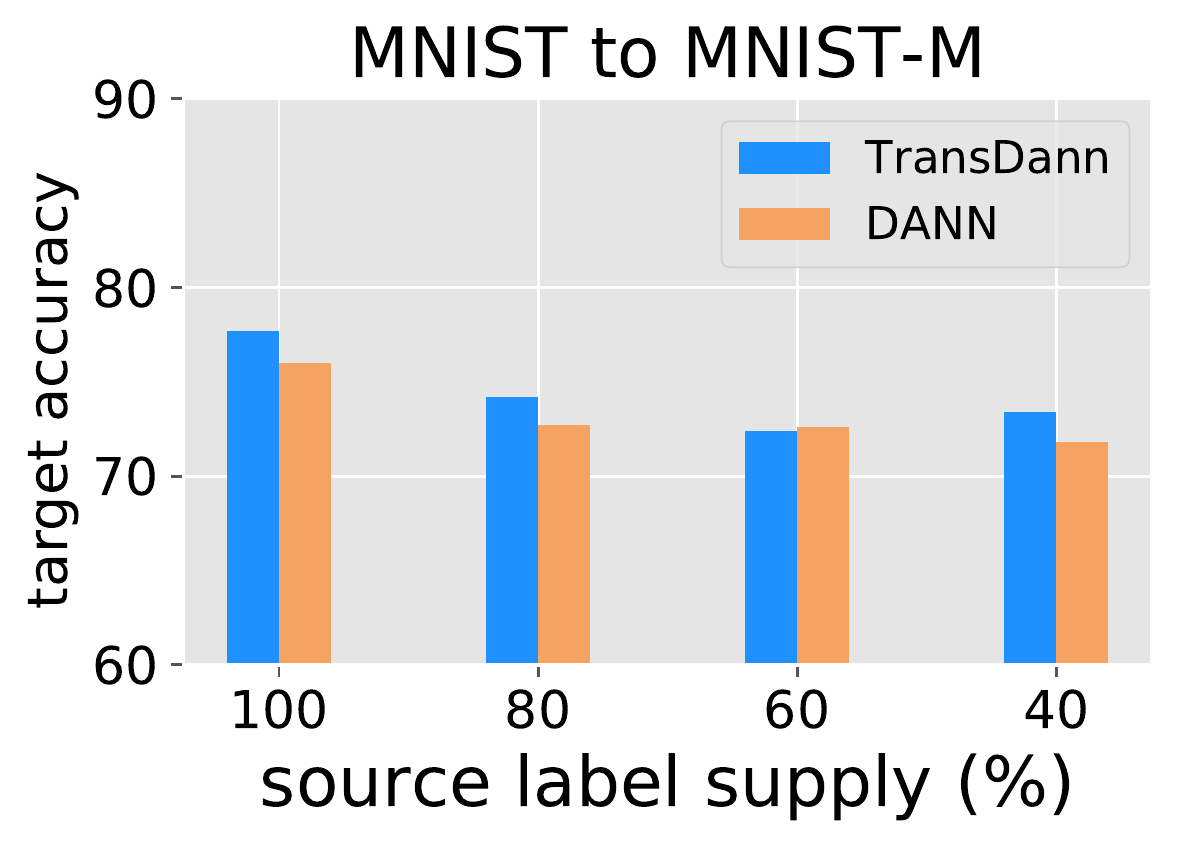}\end{subfigure}%
        \vspace*{-0.15cm}
        \caption{Performance comparison for MNIST $\rightarrow$ MNIST-M dataset. Source label supply is in  
        percentage relative to 10000 examples. }
        \label{fig:MNIST}
\vspace*{-0.45cm}
\end{figure}

\noindent{\bf On Image dataset: } 
Figure \ref{fig:MNIST} captures the performance of our DA approach on MNIST to MNIST-M. Similar as above, we observe that when source (MNIST) data supply is limited DANN's performance deteriorates and {\em TranDANN} outperforms in most of the cases. 
%%%%%

Overall, both image and text dataset evaluation validates that {\em TransDANN} outperforms DANN in LS scenarios. 

%\input{conc.tex}
%%% 							Concluding Remarks
%%%%%%%%%%%%%%%%%%%%%%%%%%%%%%%%%%%%%%%%%%%%%%%%%%%%%%%%%%%%%%%%%%%%%%%%%%%%%%%   
\section{Concluding Remarks}
In this paper, we present a novel approach, called {\em TransDANN}, which fuses adversarial learning and transductive learning methods for improved DA performance. Our approach outperforms DANN -- a state-of-the-art -- especially in scenarios where supply of source label in limited.  We have provided theoretical as well as experimental justification in support of the proposed approach. The paper unveils and establishes that adversarial learning in effect reduces any DA problem into a semi-supervised learning in a space of common representation.  Moreover, it opens up several avenues for employing various suitable semi-supervised techniques with existing adversarial based DA methods.

%%%%%%%%%%%%%%%%%%
%\clearpage
{\small
\bibliographystyle{aaai}
\bibliography{deepDABib}
}

\appendix
\section{Appendix}

\section*{Summary Statistics of Amazon Review Dataset}
Table \ref{t:dataset} (in this supplementary document) provides a detailed summary of the Amazon review dataset that were used in our experiments.
\definecolor{Gray}{gray}{0.9}
 \begin{table}[!h]
\begin{center}
\setlength\tabcolsep{0.1cm}
\begin{tabular}{@{}lcccccc@{}}
\toprule
{\textbf{Dataset}}&{Train} &{Dev.}&{Test}&{Unlab.}&{Avg.L.} &{Vocab}\\
\midrule
Books 	&	1400    & 200  & 400 &  2000 & 159 & 62k  \\
Electronics (elec) &	1398    & 200  & 400 &  2000 & 101 & 30k  \\
DVD		&	1300    & 200  & 400 &  2000 & 173 & 69k  \\
Kitchen	(kit) &	1400    & 200  & 400 &  2000 & 89 & 28k  \\

Apparel (app)	&	1400    & 200  & 400 &  2000 & 57 & 21k  \\
Camera	(cam)&	1397    & 200  & 400 &  2000 & 130 & 26k  \\
Health	(heal)&	1400    & 200  & 400 &  2000 & 81 & 26k  \\
Music	&	1400    & 200  & 400 &  2000 & 136 & 60k  \\

%Toys 		&	1400    & 200  & 400 &  2000 & 90 & 28k  \\
%Video	&	1400    & 200  & 400 &  2000 & 156 & 57k  \\
Baby		&	1300    & 200  & 400 &  2000 & 104 & 26k  \\
%Magazine	&	1370    & 200  & 400 &  2000 & 117 & 30k  \\

%Soft. 		&	1315    & 200  & 400 &  500 & 129 & 26k  \\
Sports	&	1315    & 200  & 400 &  2000 & 94 & 30k  \\
%IMDB 	&	1400    & 200  & 400 &  2000 & 269 & 44k  \\
%MR.		&	1400    & 200  & 400 &  2000 & 21 & 12k  \\
%
\bottomrule
\end{tabular}
%\vspace{-3mm}
\end{center}
\vspace{-1mm}
\caption{Amazon Review Dataset Statistics. Columns --  training, development, and test sets sizes, \# unlabeled examples, average sentence length, vocabulary size.}
\label{t:dataset}
\vspace{-5mm}
 \end{table} 
%%%%%%%%%%%%%%%%%%%%%%%%%%%%%%%%%%%%%%%%%%%%%%%%%%%%%%

%%%%%%%%%%%%%%%%%%%%%%%%%%%%%%%%%%%%%%%%%%%%%%%%%%%%%%

\section*{Elaboration of Table 2}
Table 2 in the main paper depicts the maximum \% improvement in accuracy of TransDANN over DANN, where max is computed over varying amount of labeled data. In what follows, we have presented the corresponding actual accuracy numbers for both DANN and TransDANN (for varying levels of labeled data). In all these tables (Table \ref{t:res_dann_full_100} -- \ref{t:res_dann_full_80} in this supplementary document), the rows correspond to the source domains and the columns correspond to the target domains. 

%%%%%%%%%%%%%

\setlength{\tabcolsep}{0.25em}
%%%%%%%%%%%%%%%%%%%%%%%%%%%%%%%%%%%%%%%%%%%%%%%%%%%%%%%%%%%%%%%%%%%
\begin{table*}[ht!]
\footnotesize
%\centering
\parbox{0.4\textwidth}{
%\begin{adjustbox}{angle=90}
\begin{tabular}{lcccccccccc}
\toprule  
{}&  dvd &  books &  elect &  baby &  kit &  music &  sports &  app &  cam &  health \\
%{Source} &   &&   &   &  &  & &   &   &  \\
\midrule 
\rowcolor{Gray}
dvd    &      &  77.1 &  73.9 & 76.6 & 76.2 &  79.3 &   71.1 & 83.5 & 71.5 &   56.8 \\
books  & 85.7 &       &  78.3 & 73.7 & 78.6 &  80.2 &   80.2 & 81.8 & 82.7 &   74.0 \\
\rowcolor{Gray}
elect  & 73.9 &  73.0 &       & 72.6 & 79.0 &  73.0 &   79.8 & 83.6 & 74.0 &   76.4 \\
baby   & 59.6 &  70.6 &  75.5 &      & 81.8 &  71.3 &   79.0 & 79.8 & 77.1 &   54.6 \\
\rowcolor{Gray}
kit    & 70.4 &  68.3 &  78.5 & 76.2 &      &  64.8 &   80.7 & 83.7 & 75.8 &   78.8 \\
music  & 81.4 &  77.4 &  74.0 & 75.1 & 76.2 &       &   79.6 & 80.3 & 80.2 &   74.9 \\
\rowcolor{Gray}
sports & 75.0 &  72.0 &  80.1 & 81.9 & 84.3 &  74.9 &        & 86.1 & 82.6 &   82.6 \\
app    & 72.6 &  73.2 &  79.9 & 79.9 & 80.6 &  74.7 &   76.0 &      & 77.3 &   78.8 \\
\rowcolor{Gray}
cam    & 76.0 &  70.2 &  79.9 & 78.4 & 80.5 &  73.8 &   81.5 & 75.5 &      &   81.3 \\
health & 72.1 &  71.8 &  82.6 & 82.7 & 82.9 &  77.8 &   80.9 & 84.4 & 80.4 &        \\
\bottomrule
\end{tabular}
\caption{Baseline (DANN) performance on held-out set in the target domain.  Labeled data =100\%}
%Target domain test accuracy of DANN models trained on various source-target domains on Amazon review dataset at 100 \% source label supply. 
%
\label{t:res_dann_full_100}
}
\hspace*{2.3cm}
\parbox{0.4\textwidth}{
%%\begin{adjustbox}{angle=90}
\begin{tabular}{lllllllllll}
\toprule
{} &  dvd & books & elect & baby &  kit & music & sports &  app &  cam & health \\
\midrule
\rowcolor{Gray}
dvd    &      &  75.0 &  74.2 & 77.0 & 75.8 &  79.1 &   62.5 & 82.6 & 71.5 &   55.3 \\
books  & 83.2 &       &  78.9 & 74.6 & 78.9 &  80.3 &   79.7 & 81.4 & 81.4 &   73.0 \\
\rowcolor{Gray}
elect  & 74.4 &  74.2 &       & 79.1 & 79.3 &  73.4 &   81.6 & 82.6 & 74.6 &   77.7 \\
baby   & 69.7 &  71.3 &  80.9 &      & 82.4 &  71.7 &   79.1 & 79.5 & 78.5 &   54.5 \\
\rowcolor{Gray}
kit    & 70.7 &  69.5 &  77.1 & 77.1 &      &  63.9 &   80.5 & 82.2 & 76.4 &   80.5 \\
music  & 80.5 &  78.9 &  75.4 & 74.2 & 74.8 &       &   79.5 & 80.3 & 79.9 &   74.2 \\
\rowcolor{Gray}
sports & 72.1 &  69.7 &  80.9 & 82.2 & 85.4 &  74.2 &        & 85.4 & 82.2 &   81.4 \\
app    & 72.9 &  69.9 &  79.5 & 80.5 & 79.5 &  72.9 &   73.8 &      & 76.4 &   78.5 \\
\rowcolor{Gray}
cam    & 76.8 &  71.3 &  80.3 & 74.4 & 79.5 &  71.5 &   80.1 & 78.1 &      &   80.9 \\
health & 70.3 &  71.9 &  82.2 & 83.0 & 81.4 &  77.3 &   80.7 & 84.6 & 80.9 &        \\
\bottomrule
\end{tabular}
\caption{TransDANN performance on held-out set in the target domain.  Labeled data =100\%}
%\caption*{ Target domain test accuracy of TransDANN models trained on various source-target domains on Amazon review dataset at 100 \% source label supply. }
%\label{t:res_dann_full_80}
}
\end{table*}
%%%%%%%%%%%%%%%%%%%%%%%%%%%%%%%%%%%%%%%%%%%%%%%%%%%%%%%%%%%%%%%%%%%%
%%%%%%%%%%%%%%%%%%%%%%%%%%%%%%%%%%%%%%%%%%%%%%%%%%%%%%%%%%%%%%%%%%%%
\begin{table*}[h!]
\footnotesize
%\centering
\parbox{0.4\textwidth}{
%\begin{adjustbox}{angle=90}
\begin{tabular}{lcccccccccc}\toprule
{} &  dvd & books & elect & baby &  kit & music & sports &  app &  cam & health \\
%\midrule
\rowcolor{Gray}
dvd    &      &  75.5 &  67.5 & 76.7 & 76.0 &  78.8 &   73.5 & 82.7 & 70.2 &   56.2 \\
books  & 82.9 &       &  78.8 & 73.4 & 79.3 &  79.7 &   80.7 & 82.6 & 79.6 &   75.0 \\
\rowcolor{Gray}
elect  & 75.6 &  74.4 &       & 78.3 & 79.4 &  73.8 &   80.3 & 83.8 & 78.0 &   78.2 \\
baby   & 62.0 &  71.3 &  73.9 &      & 82.6 &  63.6 &   79.1 & 81.6 & 76.4 &   63.7 \\
\rowcolor{Gray}
kit    & 69.9 &  67.6 &  79.0 & 77.9 &      &  64.6 &   80.5 & 79.3 & 76.4 &   79.5 \\
music  & 81.3 &  78.0 &  72.7 & 74.9 & 73.4 &       &   79.8 & 79.2 & 79.4 &   73.2 \\
\rowcolor{Gray}
sports & 73.6 &  72.9 &  80.1 & 81.6 & 85.4 &  74.7 &        & 84.9 & 54.6 &   81.1 \\
app    & 70.8 &  71.3 &  79.6 & 79.3 & 80.3 &  73.8 &   77.0 &      & 71.5 &   75.3 \\
\rowcolor{Gray}
cam    & 76.2 &  73.6 &  80.3 & 80.2 & 78.1 &  72.1 &   82.0 & 77.0 &      &   81.7 \\
health & 71.2 &  73.0 &  80.3 & 82.6 & 83.3 &  77.7 &   79.7 & 84.7 & 79.1 &        \\
\bottomrule
\end{tabular}
\caption{ Baseline (DANN) performance on held-out set in the target domain.  Labeled data =90\%}
%\label{t:res_dann_full_90}
}
\hspace*{2.3cm}
\parbox{0.4\textwidth}{
%\begin{adjustbox}{angle=90}
\begin{tabular}{lllllllllll}
\toprule
{} &  dvd & books & elect & baby &  kit & music & sports &  app &  cam & health \\
%\midrule
\rowcolor{Gray}
dvd    &      &  77.0 &  69.3 & 76.4 & 75.6 &  77.0 &   73.8 & 82.2 & 64.6 &   58.8 \\
books  & 83.0 &       &  79.3 & 73.8 & 79.1 &  79.1 &   80.7 & 81.4 & 82.2 &   74.0 \\
\rowcolor{Gray}
elect  & 76.2 &  75.0 &       & 76.6 & 80.5 &  77.1 &   80.3 & 82.2 & 82.0 &   78.9 \\
baby   & 71.9 &  70.7 &  79.3 &      & 83.0 &  73.2 &   80.1 & 83.8 & 74.4 &   69.3 \\
\rowcolor{Gray}
kit    & 69.1 &  66.0 &  78.3 & 80.1 &      &  67.8 &   80.1 & 83.4 & 76.8 &   79.7 \\
music  & 80.7 &  77.0 &  71.7 & 73.6 & 75.4 &       &   81.4 & 78.3 & 78.1 &   72.5 \\
\rowcolor{Gray}
sports & 74.6 &  73.2 &  81.2 & 82.4 & 84.8 &  75.2 &        & 84.6 & 80.7 &   80.9 \\
app    & 67.8 &  74.0 &  78.7 & 77.3 & 82.2 &  75.2 &   76.8 &      & 69.7 &   77.9 \\
\rowcolor{Gray}
cam    & 78.1 &  72.9 &  81.4 & 78.9 & 78.5 &  68.2 &   82.6 & 74.0 &      &   83.0 \\
health & 69.9 &  73.2 &  78.9 & 82.8 & 84.0 &  77.5 &   80.7 & 85.9 & 77.1 &        \\
\bottomrule
\end{tabular}
\caption{TransDANN performance on held-out set in the target domain.  Labeled data =90\%}
%\label{t:res_dann_full_80}
}
\end{table*}
%%%%%%%%%%%%%%%%%%%%%%%%%%%%%%%%%%%%%%%%%%%%%%%%%%%%%%%%%%%%%%%%%%%%
\begin{table*}[h!]
\footnotesize
%\centering
\parbox{0.4\textwidth}{
%\begin{adjustbox}{angle=90}
\begin{tabular}{lllllllllll}
\toprule
{} &  dvd & books & elect & baby &  kit & music & sports &  app &  cam & health \\
%\midrule
\rowcolor{Gray}
dvd    &      &  76.7 &  68.4 & 75.7 & 76.2 &  75.2 &   66.7 & 79.6 & 68.9 &   59.2 \\
books  & 82.3 &       &  79.3 & 73.3 & 79.0 &  79.4 &   79.9 & 80.4 & 80.2 &   75.5 \\
\rowcolor{Gray}
elect  & 70.8 &  75.6 &       & 72.3 & 71.0 &  67.1 &   80.8 & 82.9 & 77.5 &   75.1 \\
baby   & 61.8 &  69.8 &  74.9 &      & 79.2 &  72.5 &   78.6 & 73.4 & 74.7 &   68.7 \\
\rowcolor{Gray}
kit    & 63.2 &  68.8 &  78.3 & 79.3 &      &  69.3 &   80.0 & 76.4 & 76.0 &   79.8 \\
music  & 82.7 &  78.3 &  73.6 & 74.2 & 74.2 &       &   81.1 & 79.7 & 78.9 &   73.0 \\
\rowcolor{Gray}
sports & 71.9 &  70.6 &  80.6 & 78.7 & 82.7 &  74.6 &        & 84.2 & 75.3 &   80.6 \\
app    & 69.4 &  73.6 &  78.9 & 78.3 & 82.0 &  75.9 &   77.9 &      & 70.4 &   76.8 \\
\rowcolor{Gray}
cam    & 75.0 &  72.1 &  78.9 & 78.9 & 76.3 &  69.8 &   80.9 & 68.8 &      &   82.1 \\
health & 71.0 &  74.0 &  81.2 & 80.6 & 83.1 &  78.3 &   78.4 & 84.6 & 78.2 &        \\
\bottomrule
\end{tabular}

\caption{Baseline (DANN) performance on held-out set in the target domain.  Labeled data =80\%}
%\label{t:res_dann_full_80}
}
\hspace*{2.3cm}
\parbox{0.4\textwidth}{
%\begin{adjustbox}{angle=90}
\begin{tabular}{lllllllllll}
\toprule
{} &  dvd & books & elect & baby &  kit & music & sports &  app &  cam & health \\
\midrule
\rowcolor{Gray}
dvd    &      &  78.3 &  72.1 & 72.5 & 76.6 &  77.7 &   72.5 & 78.3 & 66.2 &   61.3 \\
books  & 81.6 &       &  78.5 & 74.8 & 79.3 &  79.5 &   78.5 & 77.5 & 78.5 &   75.6 \\
\rowcolor{Gray}
elect  & 61.9 &  77.5 &       & 65.4 & 51.6 &  76.6 &   81.2 & 81.8 & 75.0 &   68.2 \\
baby   &  NaN &  70.7 &  71.5 &      & 80.5 &  70.7 &   80.1 & 57.0 & 76.0 &   54.5 \\
\rowcolor{Gray}
kit    & 50.6 &  68.2 &  79.1 & 79.7 &      &  71.7 &   81.1 & 78.1 & 76.8 &   80.5 \\
music  & 82.0 &  80.1 &  76.4 & 74.0 & 76.6 &       &   82.0 & 79.5 & 80.7 &   74.2 \\
\rowcolor{Gray}
sports & 71.5 &  70.1 &  79.3 & 77.3 & 80.1 &  73.2 &        & 84.4 & 77.1 &   80.1 \\
app    & 73.0 &  72.9 &  78.5 & 78.9 & 82.0 &  74.2 &   76.8 &      & 69.7 &   75.8 \\
\rowcolor{Gray}
cam    & 74.8 &  72.3 &  79.3 & 78.7 & 77.0 &  66.8 &   77.3 & 57.0 &      &   79.9 \\
health & 70.3 &  74.0 &  80.3 & 79.5 & 84.0 &  77.9 &   78.7 & 85.7 & 77.0 &        \\
\bottomrule
\end{tabular}
\caption{TransDANN performance on held-out set in the target domain.  Labeled data =80\%}
\label{t:res_dann_full_80}
}
\end{table*}

%%%%%%%%%%%%%%%%%%%%%%%%%%%%%%%%%%%%%%%%%%%%%%%%%%%%%%%%%%%%%%%%%%%%%%%%%%%
\end{document}